\documentclass[twoside]{IEEEtran}

\usepackage[T1]{fontenc}
\usepackage[utf8]{inputenc}
\usepackage{amsmath}
\usepackage{amsthm}
\usepackage{amsfonts}
\usepackage{textpos}
\usepackage{float}
\usepackage{multirow}
\usepackage{tabulary}
\usepackage{prettyref}
\usepackage{titleref}
\usepackage{fancyhdr}
\usepackage[final]{pdfpages}
\usepackage{hyperref}
\usepackage{subfig}
\usepackage{morefloats}
\usepackage{natbib}

\usepackage{graphicx}
\usepackage[rel]{overpic}
\def\TReg{\textsuperscript{\textregistered}}

\def\modified#1{#1}

\title{Unsupervised Construction of Human Body Models Using Principles of Organic Computing}

\author{\emph{Thomas~Walther and Rolf P. W\"urtz}\\[5mm]
Department of Electrical Engineering and Information Technology and \\
Institute for Neural Computation, Ruhr-University Bochum, Germany\\
thomas.walther@rub.de, rolf.wuertz@ini.rub.de}

\begin {document}
\maketitle
\begin{abstract}
Unsupervised learning of a generalizable model of
the visual appearance of humans from video data is of major
importance for computing systems interacting naturally with
their users and others.
We propose a step towards automatic behavior understanding
by integrating principles of Organic Computing into the posture
estimation cycle, thereby relegating the need for human
intervention while simultaneously raising the level of system
autonomy. The system extracts coherent motion from moving
upper bodies and autonomously decides about limbs and their
possible spatial relationships. The models from many videos are
integrated into meta-models, which show good generalization
to different individuals, backgrounds, and attire. These models
allow robust interpretation of single video frames without 
temporal continuity and posture mimicking by an android robot.
\end{abstract}

\pagestyle{fancy}
\fancyhead{}
\fancyfoot{}
\fancyhead[RO,LE]{\thepage}
\fancyhead[RE]{Walther and W\"urtz}
\fancyhead[LO]{Unsupervised Construction of Human Body Models Using Principles of Organic Computing}
\thispagestyle{empty}

\section{Introduction}

Humans show unmatched expertise in analyzing and interpreting body
movement and anticipating intentions and future behavior. This skill
of social perception is one of the foundations of effective and smooth
interaction of humans inhabiting a complex environment.  The benefits
of machines capable of interpreting human motion would be enormous:
applications in health care, surveillance, industry and
sports~\citep{Gavrila99,Moeslund06} promise a broad market.  Despite
significant effort~\citep{Poppe07} to transfer human abilities in
motion estimation and behavioral interpretation to synthetic systems,
automatically \emph{looking at people}~\citep{Gavrila99} remains among
the `most difficult recognition problem[s] in computer
vision'~\citep{Malik04} there is no technical solution matching human
competency in vision-based motion capturing (VBMC).

Artificial VBMC systems should be enhanced by learning lessons from
human perception. To that end, we have pursued an \emph{organic
  computing} (OC)~\citep{Wuertz08} approach and introduce cerebral
information processing paradigms (including \emph{self-organization},
\emph{self-optimization}, \emph{concept building, generalization}, and
\emph{non-trivial learning}~\citep{Poggio04} into vision-based human
posture analysis. The resulting system is able to acquire conceptual
models of the upper human body in a completely autonomous manner: the
learning procedures are based on only a few general principles. This
strategy significantly reduces human workload and allows
self-optimization of the generated models. While autonomous model
learning and knowledge agglomeration take place in simple scenarios,
the conceptual nature of the retrieved body representations allows for
generalization to more complex scenarios and holds opportunities for
model \emph{adaptation and enhancement loops}, which might perform
continuous, non-trivial learning as found in the human brain.

\begin{figure*}[p]	
\centering
\begin{overpic}[height=0.95\textheight]{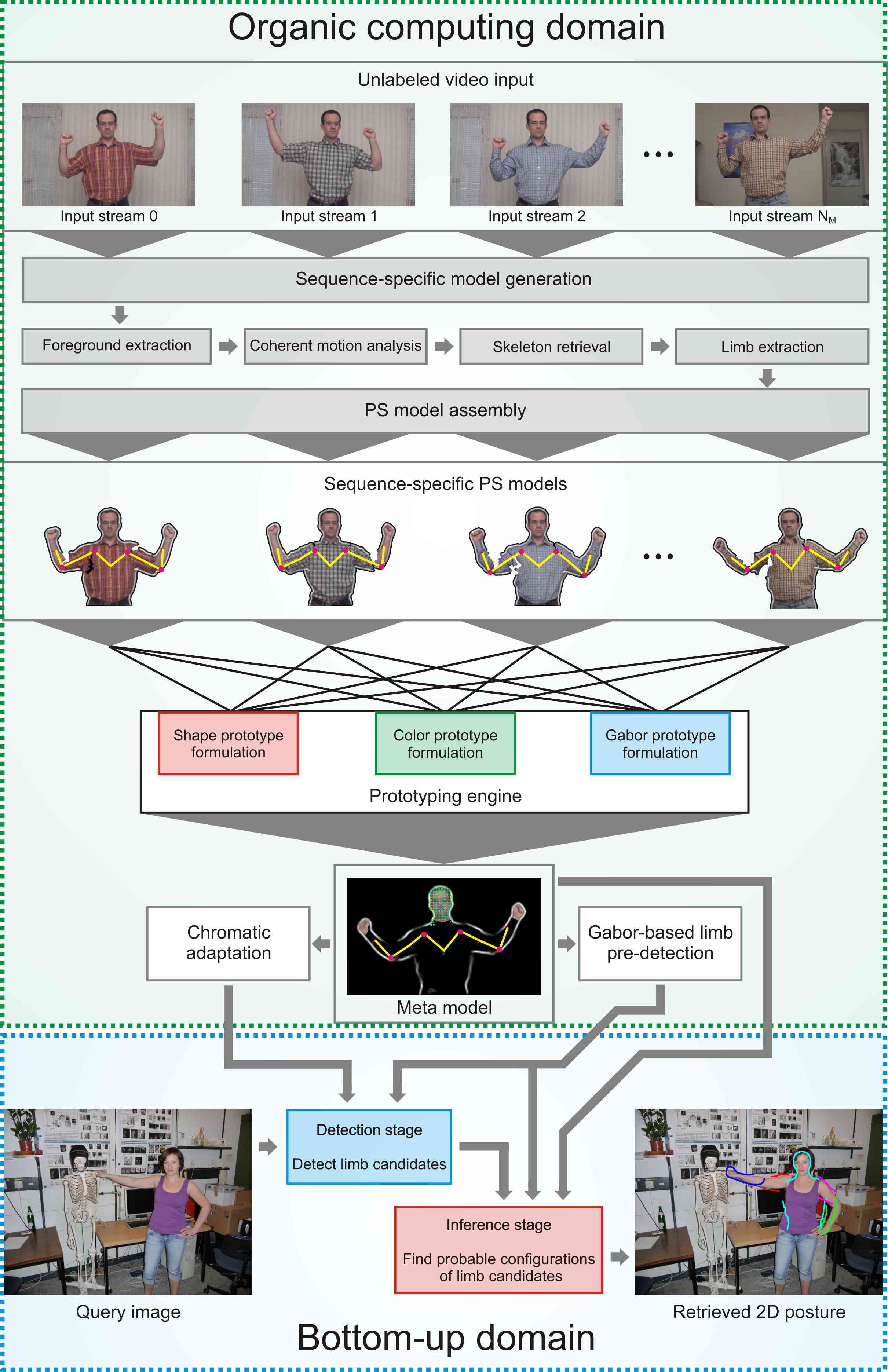}
\put(50,97.2){\Large\sf (a)}
\put(44,79.2){\sf (b)}
\put(14.5,74.5){\scriptsize\sf (c)}
\put(31,74.5){\scriptsize\sf (d)}
\put(45,74.5){\scriptsize\sf (e)}
\put(61,74.5){\scriptsize\sf (f)}
\put(39,69.6){\sf (g)}
\put(42,64.6){\sf (h)}
\put(20,44.8){\scriptsize\sf (i)}
\put(36,44.8){\scriptsize\sf (j)}
\put(51.5,44.8){\scriptsize\sf (k)}
\put(20,30.1){\sf (l)}
\put(36.5,26.2){\sf (m)}
\put(52,30.1){\sf (n)}
\put(33,14.8){\scriptsize\sf (o)}
\put(42,6.7){\scriptsize\sf (p)}
\put(45,1.9){\Large\sf (q)}
\end{overpic}
	\caption[System overview]{Schematic overview of the proposed
          VBMC system: green regions indicate organic learning
          components, blue areas envelop standard bottom-up VBMC
          (cf. section~\ref{section:2}) components.}
	\label{fig:SystemOverview}
\end{figure*}
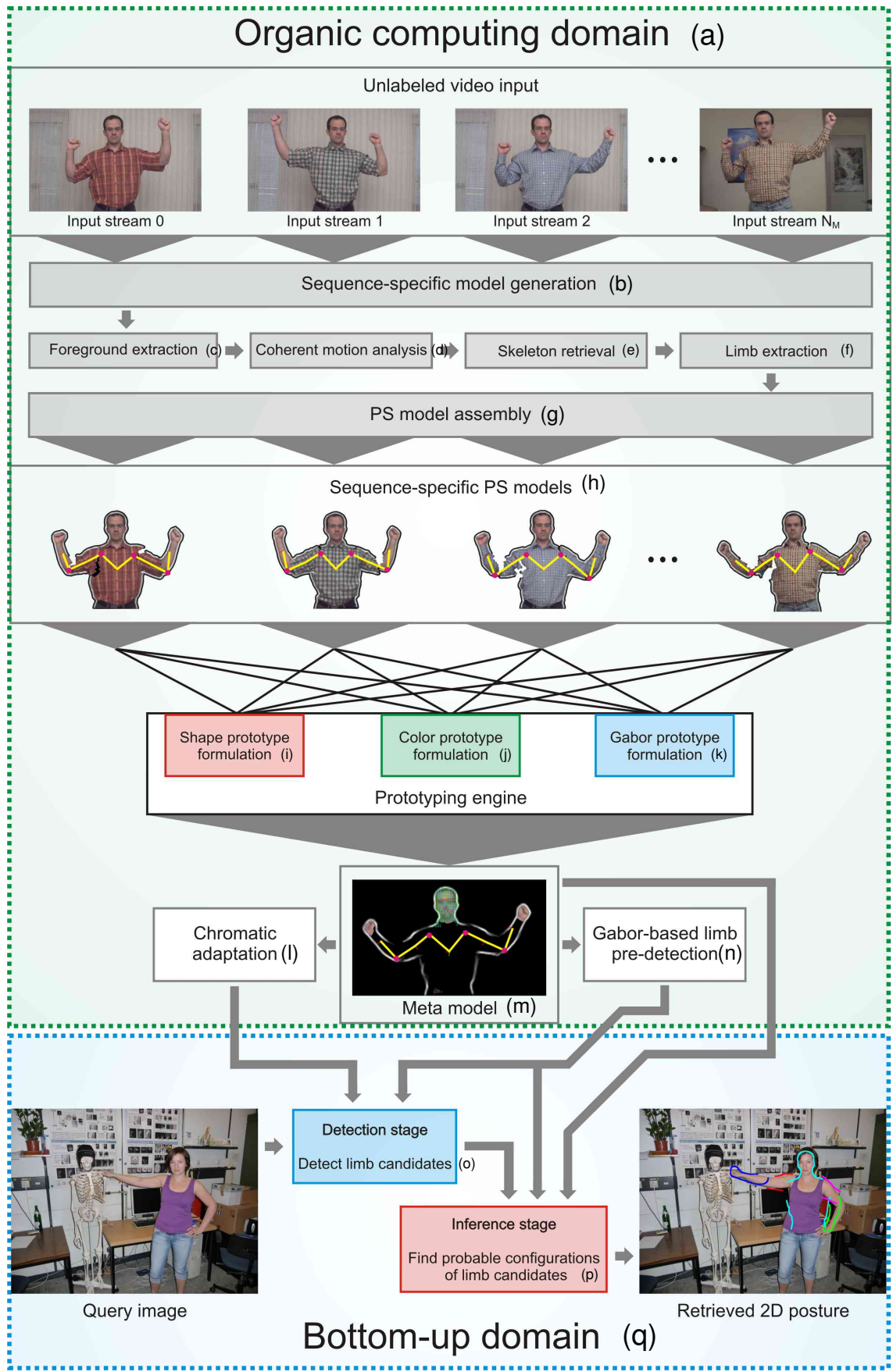

\modified{Fig.~\ref{fig:SystemOverview} provides a schematic overview
  of the system and is referred to throughout the paper for all
  components.  In section~\ref{section:2} we give an overview of
  VBMC approaches that have been considered or used in this work and
  discuss their strenghts and weaknesses. Section~\ref{section:3}
  describes the details of learning a body model from a single video
  of human motion. In section~\ref{section:4} many such models are
  combined into a meta-model, which captures the invariant cues of the
  single models. Section 5 tests the learned meta-model on still images
  with different backgrounds, individuals, attire, etc. Relations with
  powerful supervised models are discussed. The paper closes with a short
  discussion.
}

\section{\modified{Previous work} in vision-based human motion capturing} \label{section:2}

Following~\cite{Poppe07}, VBMC methods can be classified
into \emph{model-based}, \emph{generative} approaches and
\emph{model-free}, \emph{discriminative} methods
(cf. also~\citep{Cipolla06}). Model-based schemes incorporate
\emph{top-down} and \emph{bottom-up} techniques, while the model-free
domain employs \emph{learning-based} and \emph{exemplar-based} pose
estimation.

To stay in scope, we leave an in-depth discussion of top-down and
discriminative techniques to~\cite{Poppe07}
or~\cite{Walther11}. Bottom-up solutions form an important mainstay of
our own approach and are thus investigated more closely. Nevertheless,
our focus is on autonomous, fully unsupervised VBMC strategies that
try to enhance systemic autonomy and directly relate to the
organically inspired posture estimation method proposed in this paper.

\subsection {Bottom-up posture estimation}

A generic bottom-up (or \emph{combinatorial}~\citep{Ricketts07})
posture estimation system follows the principle formulated
in~\citep{Black06a}: `measure locally, reason globally.'  Local
measurement treats the human body as an ensemble of
`quasi-independent'~\citep{Isard03} limbs, which much alleviates the
complex model coupling inherent in top-down approaches. Imposing
independence, `image measurements'~\citep{Isard03} of single limbs can
be performed separately by a dedicated \emph{limb detector}
(LD)~\citep{Black06b,Zisserman07}, which moves the burden of matching a
given body part model to some well-chosen \emph{image
  descriptors}~\citep{Metaxas07,Poppe07}. The selection of appropriate
images descriptors as well as construction and application of LDs
require domain knowledge of and concept building by human supervisors.
For many object categories, histograms of oriented gradient (HOG) features
seem to be a good choice, allowing object classification by linear
discriminant analysis~\citep{hariharan-who-2012}.

To organize the data from local measurements, pending inter-limb
dependencies come into play during global reasoning.
`Assemblies'~\citep{Moeslund06} of detector responses are retrieved
that comply well with kinematically meaningful human body
configurations. The majority of bottom-up systems employ
\emph{graphical models}~\citep{Black06a} (GMs) to encode human body
assemblies: each node in the model's graph structure correlates to a
dedicated body part, whereas the graph's edges encode (mostly)
`spring-like'~\citep{Isard03,Lan05} kinematic relationships between
single limbs.

Using GMs for global inference, a configuration becomes more
`human-like'~\citep{Huttenlocher00} if all LDs return low matching cost
and the `springs' between the body parts are close to their resting
positions. This can conveniently be formulated by means of an
\emph{energy functional}, whose global minimum represents the most
probable posture of the captured subject.  However, minimization for
arbitrary graphs and energy functions is
NP-hard~\citep{Huttenlocher05}. Thus,~\cite{Huttenlocher00} propose to
restrict the graphs to be \emph{tree-like} and further restrictions on
the energy function to allow for computationally feasible posture
inference using \emph{dynamic programming}~\citep{Huttenlocher05}.  We
follow this approach by boosting the \emph{pictorial
  structure}~\citep{Fischler73} (PS) approach
of~\cite{Huttenlocher00,Huttenlocher05} with organic components to
maximize system autonomy.


\begin{figure*}[t]
	\centering
\subfloat[\label{fig:Frame30}]{\includegraphics[width=0.32\textwidth]{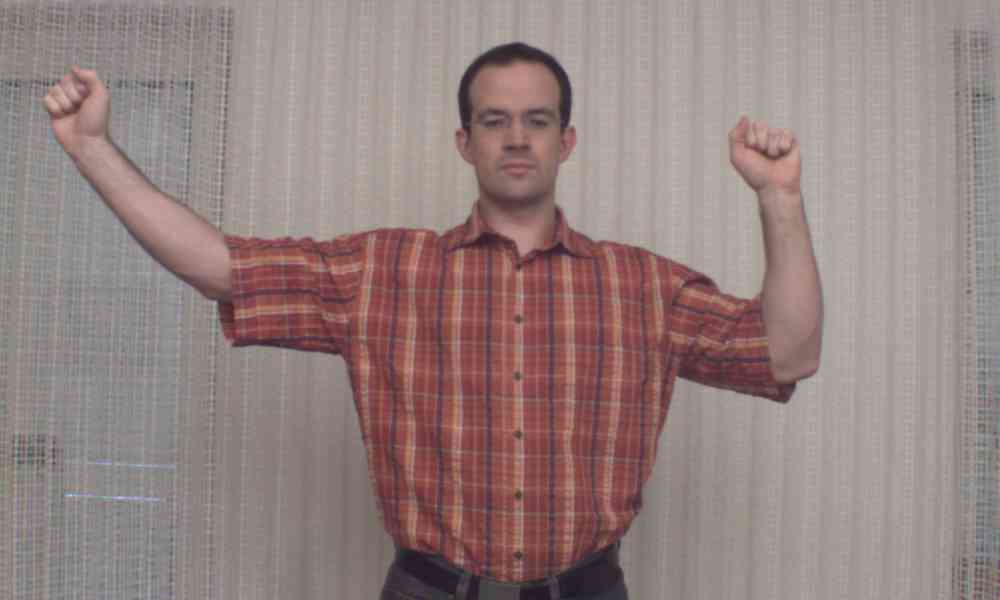}}
	\hspace{0.1cm}
	\subfloat[\label{fig:Frame115}]{\includegraphics[width=0.32\textwidth]{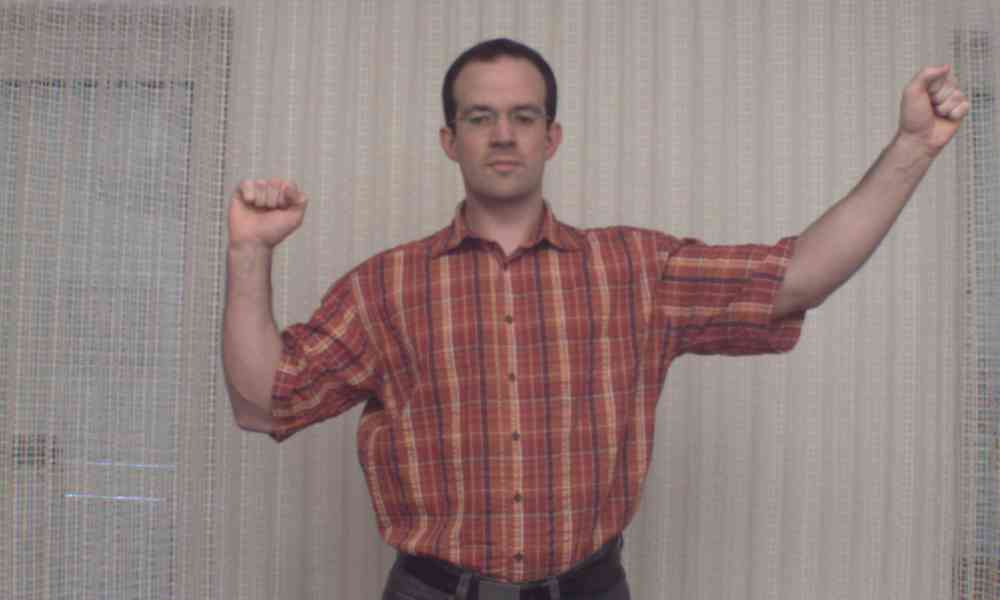}}
	\hspace{0.1cm}
	\subfloat[\label{fig:Frame147}]{\includegraphics[width=0.32\textwidth]{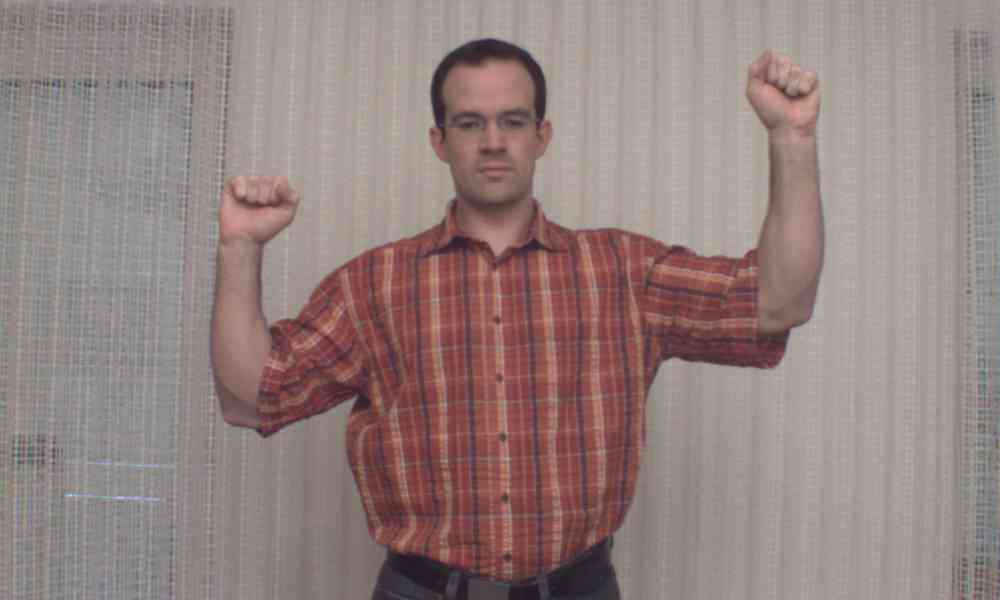}}\\
\subfloat[\label{fig:ReferenceFrame}]{\includegraphics[width=0.32\textwidth]{Images/fig_StandardTrainingSequence/Sequenz006A/Sequenz147.jpg}}
\hspace{0.1cm}
\subfloat[\label{fig:GraphCutResult}]{\includegraphics[width=0.32\textwidth]{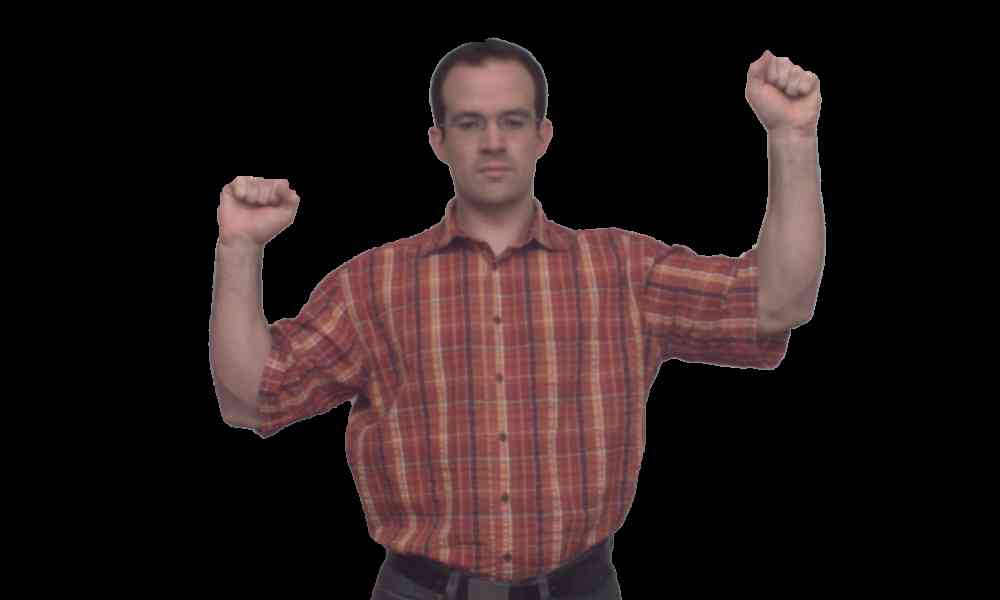}}\hspace{0.1cm}
\subfloat[\label{fig:FinalFeaturePlacement}]{\includegraphics[width=0.32\textwidth]{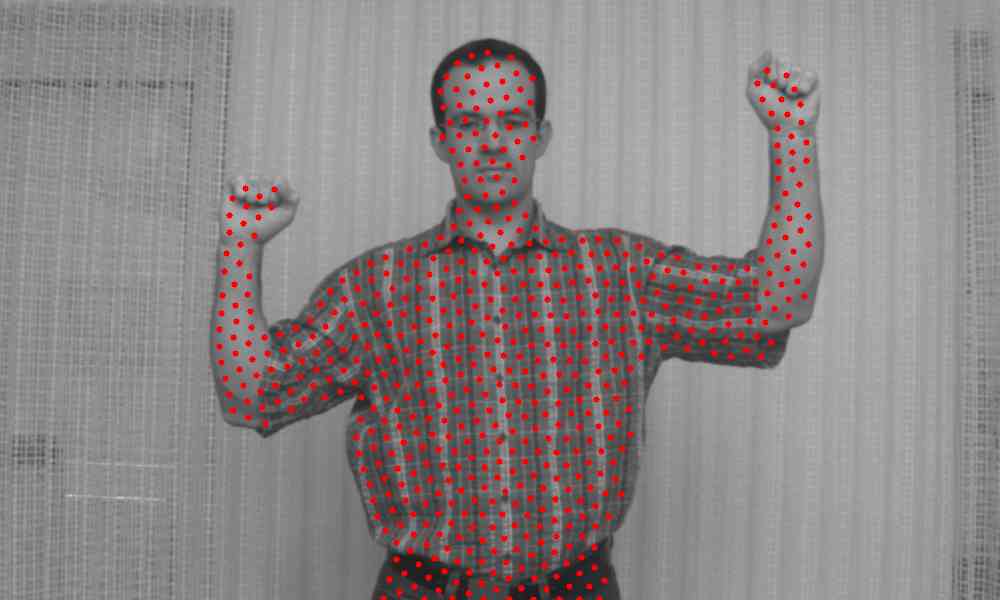}}
\\ \subfloat[\label{fig:FeatureGroups}]{\includegraphics[width=0.32\textwidth]{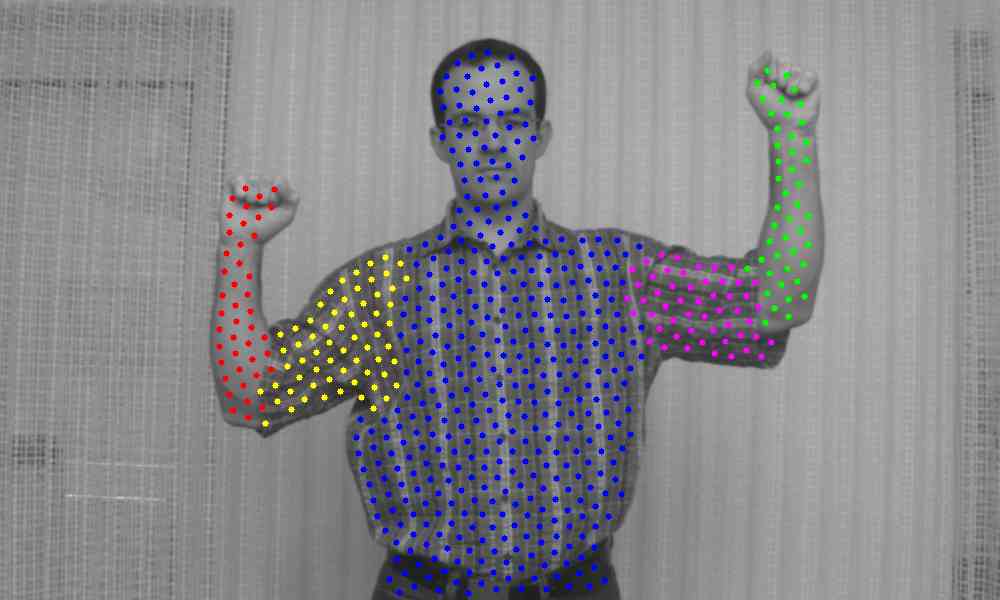}}\hspace{0.1cm}
\subfloat[\label{fig:Skeleton}]{\includegraphics[width=0.32\textwidth]{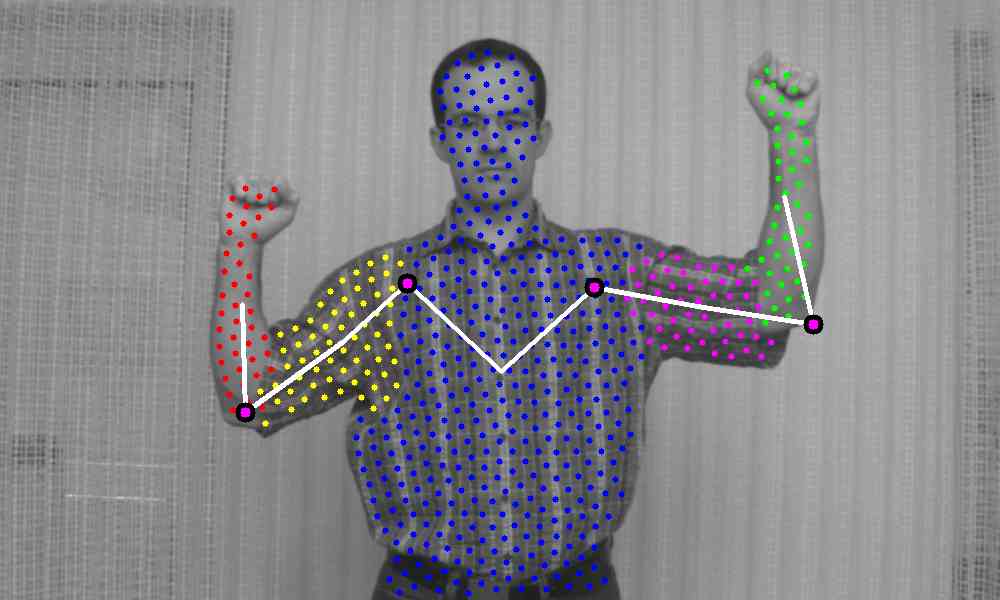}}
\hspace{0.1cm}
\subfloat[\label{fig:LimbSegmentation}]{\includegraphics[width=0.32\textwidth]{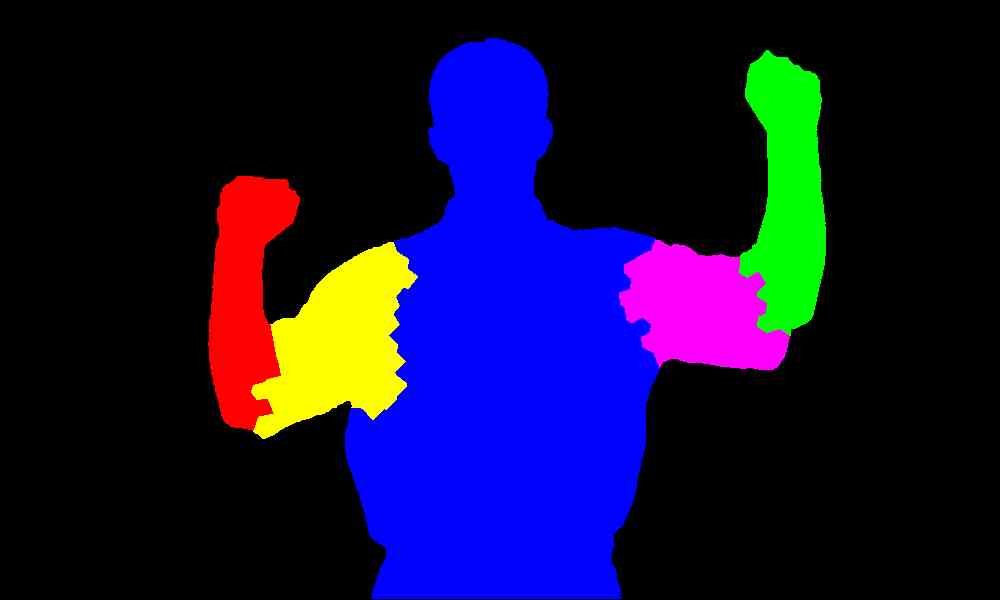}}
	\caption[]{Frames 30~(Fig.~\ref{fig:Frame30}),
            115~(Fig.~\ref{fig:Frame115}), and 147~(Fig.~\ref{fig:Frame147}) from a
            standard scenario. Fig.~\ref{fig:ReferenceFrame} shows  the reference
            frame from that sequence. Graph cut results are given
            in~Fig.~\ref{fig:GraphCutResult}, the resulting feature
            placement is sketched
            in~Fig.~\ref{fig:FinalFeaturePlacement}. Fig.~\ref{fig:FeatureGroups}
            shows the trajectory segmentation for the reference frame,
            Fig.~\ref{fig:Skeleton} the corresponding
            skeleton, and Fig.~\ref{fig:LimbSegmentation} the resulting
            limb template masks.}
\end{figure*}

\subsection{Autonomous VBMC}
We now  discuss a novel generation of \emph{autonomous} VBMC
methods --- these systems learn body models autonomously
from unlabeled training input and tend to show fair generalization
performance while applying the learned patterns to novel
scenarios. We discuss a selection of recent, autonomous VBMC
approaches; as the techniques come closer to the focus
of this paper we step into more detail.

Inspired by point-light display experiments~\citep{Song03} learns
\emph{decomposable triangulated graph} (DTG) models of the human
body. DTG nodes correspond to single limbs, edges express mutual
dependencies. Implying that limbs far apart in the kinematic chain
have no significant influence on each other, it seems reasonable to
postulate conditional independence of the triangle
cliques~\citep{Song03} in the DTG, which leads to efficient mechanisms
for human detection/posture retrieval in scenarios of increasing
complexity. Tracked \emph{Kanade-Lucas-Tomasi} (KLT)~\citep{Yan06}
features act as `virtual markers' (one per body part) during pose
inference, indicating temporally varying limb positions. The quality
of the automatically generated models does not reach that of
hand-crafted body representations.  Further, single KLT features are
unlikely to behave like physical markers and trace limb movements
exactly because of feature loss and background distractions.

\cite{Pollefeys08} also propose to infer body structure automatically
from point features. Representing moving limbs by a multitude of
trackers, that scheme can cope with moderate feature loss and extends
standard \emph{structure from motion} (SFM) to deal with non-rigid and
articulated structures. Their main assumption is that `trajectories
from the same motion lie in the same rigid or nonrigid motion
subspace, whereas those from different motions do not.' Each feature's
motion subspace is estimated by manually adjusted \emph{local
  sampling}, distances between feature trajectories are measured by
the \emph{principal angle}~\citep{Golub96} between the respective
subspaces.  The inter-trajectory distances form an \emph{affinity
  matrix}~\citep{Pollefeys08}, upon which recursive \emph{spectral
  clustering}~\citep{luxburg-tutorial} identifies groups of coherently
moving features, and \emph{minimum spanning tree}~\citep{Pollefeys08}
techniques based on the principal angles succeed to retrieve a
kinematic skeleton. This system has a high degree of autonomy as
almost no human supervision is required in the model learning loop,
and the quality of the body parts and the kinematic skeleton comes
close to human intuition.

\cite{Ross10} follow the same paradigm to human articulation
structure from point feature motion: a body model is set up that
includes latent structural variables describing the assignment of
tracked features to the skeletal bones and the connectivity of the
body structure.  Residual model variates identify limb feature
coordinates and locations of potential joints. The `expected complete
log-likelihood of the observed [point motion] data given the
model'~\citep{Ross10} should obviously be maximized if the model
parameters closely match the true structure of the captured
entity. This \emph{optimal} model is acquired by a combined learning
scheme: first, \emph{affinity propagation}~\citep{Frey07}
finds an initial point-to-limb assignment, temporarily neglecting
skeletal connections. Then, other model variates (excluding latent
structural connectivity) are refined, making intense use of
\emph{expectation maximization}~\citep{Dempster77} (EM). Based on these
preparations, iteration starts: joints between limbs are putatively
introduced, and the most likely joint is kept. With the updated
topology, the EM loop repeats, simultaneously performing a periodic
update of all latent feature-to-limb assignments. The model evoking
maximal complete log-likelihood is output as the optimal solution.
The system can handle articulated SFM tasks for human, animal, and
technical structures without supervision. Non-rigidity like
cloth deformation gives rise to unreasonable limb estimates/kinematic
skeletons.

Those schemes yield sparse limb representations,~\cite{Krahnstoever03} 
takes one step beyond; beginning
with sparse SFM-like techniques based on KLT feature tracking and
standard $K$-means trajectory clustering, groups of coherently moving
features are identified in fronto-parallel scenarios. The basic
clustering objective yields perceptually acceptable
approximations of the true limb structure in all performed experiments
if $K$ is properly selected manually.  The tracked features act as
seeds to an EM segmentation scheme relying on shape, color, and motion
cues that yields fleshed-out limb templates precisely encoding the
appearance of the captured subject. Based on the motion behavior of
these templates, probabilistic tree spanning techniques identify
likely joints between the extracted body parts and generate a
well-defined kinematic skeleton for the articulated
target. \cite{Krahnstoever03} successfully extracts body appearance and
topology from synthetic and real input. Except the selection of $K$, the
method is unsupervised and thus a good starting point for autonomous
learning.

Similarly, \cite{Kumar08} extract coherent motion
\emph{layers} from video footage: input frames are
first split into rectangular patches, over which a conditional random
field~\citep{Wallach04} is defined. Belief propagation then identifies
patches that follow the same rigid motion model between consecutive
frames~\citep{Kumar08}. From those coherently moving {\em motion
  components}, initial \emph{body part segments} are formed by
integrating component information from all input frames, which carry
sufficient information to seed \emph{limb refinement}~\citep{Kumar08}:
\emph{$\alpha$-expansion} and \emph{$\alpha$-$\beta$-swap}
mechanisms~\citep{Veksler99} cut out precise templates for each
limb. This scheme achieves competitive limb segmentation results that
correspond well to human intuition, while maintaining a significant
degree of autonomy. On the other hand, it requires computationally
demanding algorithms and has an unwieldy system structure. Skeleton
retrieval and non-rigidity are not discussed.

\cite{Kumar10} build upon~\citep{Kumar08} in order to learn
\emph{object category models} (OCMs) of animals. These OCMs are
encoded as \emph{layered pictorial structures} (LPSs) which can be
learned autonomously from multiple video sequences that contain a
dedicated animal category.  LPS nodes comprise sets of
category-specific {\em shape and texture exemplars} for each limb,
edges express spatial relations between the body parts.
\cite{Kumar10} use their OCMs to guide a probabilistic foreground
segmentation scheme that shows acceptable performance in cutting out
members of the encoded categories from cluttered images. This method
shows a promising capability of concept building and generalizes well
to novel situations. Using exemplar sets instead of limb prototypes,
memory requirements are likely to become an issue for larger training
databases, and accessing specific templates in large exemplar
populations might be computationally demanding. \cite{Kumar10} draw
inspiration from~\cite{Stenger04a} and organize exemplar data in a
hierarchical manner to speed up access during LPS matching.

\cite{Ramanan06} use the pictorial structure paradigm, learning
tree-like PS representations of animals from given video input:
assuming that animal limbs can roughly be described by rectangular
approximations, rectangle detectors identify candidate body parts in
all input frames. Candidates that keep up `coherent appearance across
time'~\citep{Ramanan06} are found by clustering, resulting in
`spatio-temporal tracks of limbs,' tracks violating a
predefined model of smooth motion are pruned.  The remaining clusters
are interpreted as body parts whose appearance is in LDs at the PS
nodes. Skeletal structure is derived via a `mean distance-based'
minimum spanning tree approach.  The method operates without human
guidance and displays fair generalization. It has been
further developed in~\citep{yang-ramanan-2013}, where limbs are modeled
as mixtures of undeformable parts. This achieves excellent performance
on difficult datasets like Buffy~\citep{Ferrari12} but was not considered
during development of our model.

\cite{Zisserman07} extends~\cite{Ramanan06}'s approach to humans: The
PS model is defined a priori, making it unnecessary to perform an
unreliable guess on the correct number of body parts or the desired
kinematic skeleton. Limb appearance is found by
applying~\cite{Ramanan06}'s track clustering algorithm to the first
few frames of an input stream. The resulting PS representations
generalize well to all residual frames of the sequence. Despite good
results, the weak rectangle detectors easily go astray in complex
scenarios, possibly spoiling PS pattern formulation. They are replaced
by \emph{stylized detectors} defined as a body configuration that is
`easy to detect and easy to learn appearance from'.  Both desiderata
hold for {\em lateral walking poses}, which can be detected with a
specifically tuned PS body model.  Once such a \emph{stylized pose
  pictorial structure} reliably locked on to a lateral walking pose in
some input frame, the appearance (color distribution) of each limb can
be retrieved and used to form classifiers for the single body parts,
which act as LDs in a {\em general pose pictorial structure model}
that allows to infer posture in the residual frames of the processed
sequence. From the OC perspective, the approach by~\cite{Zisserman07} holds
promise w.\,r.\,t.\ autonomous model evolution, but the evolved PS
patterns becomes `highly person-specific'~\citep{Zisserman07} and would
probably generalize poorly to differently dressed individuals. The
structure of the initial PS models is defined through human expertise.
For the case of sign language this method is enhanced
by~\cite{pfister-bmvc-2013} through correlations with mouth movements.

\section{Autonomous acquisition of human body models} 
\label{section:3}

A graphical \emph{human body model} (HBM) with well-designed limb
detectors is one mainstay of successful bottom-up human posture
estimation/human motion analysis (HPE/HMA). In the OC context,
modeling effort should be left to the machine learning model structure
and salient features autonomously from input data.  However, this
strategy is hampered by real-world phenomena like limited camera
performance, background clutter, illumination changes, occlusion
issues, etc., all of which have dramatic al impact on the model
learning process, cf.~\citep{Walther11}.

To reduce these problems we impose restrictions like a single
individual controlled illumination, and slow and smooth movement on
our learning scenarios, but include all movements supposed to be
learned. We will assume that
\begin{enumerate}
\item limbs are coherently moving subparts of the human body,
connected by a tree-like kinematic skeleton, 
\item throughout a given input sequence, the appearance of all limbs
can be assumed near constant,
\item all limbs of interest are exercised vividly in a given training
sequence.
\end{enumerate}

 Based on these fundamental rules, fully autonomous extraction of
 \emph{sequence-specific} HBMs from short input video streams of low
 visual complexity becomes viable
 (Fig.~\ref{fig:SystemOverview}~b). Three exemplary frames from an
 input sequence are sketched in
 Fig.~\ref{fig:Frame30} through~\ref{fig:Frame147}.

\subsection{Acquiring limb patterns from video input}

We begin by retrieving sequence-specific \emph{limb representations}
via \emph{coherent motion analysis} (Fig.~\ref{fig:SystemOverview}d): let
$\overline{\text{DDI}}^t(\mathbf{x})$ represent the number of
foreground \emph{active motion pixels} (AMPs) in morphologically
manipulated \emph{double difference images} (DDI)~\citep{Minoh96}
derived from a given input stream. From those, the \emph{reference
  frame} $t^*$ is defined as the one with with maximal average DDI.
The DDI-based foreground estimate in $I^{t^*}(\mathbf{x})$ yields a
rough approximation of the foreground shape (the moving subject). A
relatively sparse \emph{foreground map} and a compact, dense
\emph{background map} are constructed using the \emph{graph cut}
segmentation of~\cite{Boykov01}.  Based on the \emph{maximum
  flow} technique~\citep{Boykov04}, we employ the graph cut scheme
of~\cite{Mannan08} to perform precise foreground extraction in
$I^{t^*}(\mathbf{x})$, (Fig.~\ref{fig:GraphCutResult}). Remaining
outliers in the resulting \emph{foreground proposal map}
$P^{t^*}_\text{F}(\mathbf{x})$ are eradicated via morphological
closing (fig~\ref{fig:SystemOverview}c).

Next, the motion of the subject's upper body is traced through all
frames $t\neq t^*$ by applying
\emph{Kanade-Lucas-Tomasi}~\citep{Tomasi91} forward/backward tracking
to $N_\text{F}$ features, which are distributed isotropically on the
extracted foreground entity,
(Fig.~\ref{fig:FinalFeaturePlacement}). Inter-feature distance is
automatically tuned using a test-and-repeat scheme to achieve constant
$N_{\text{F}}$ for all input scenarios.

From the so generated \emph{feature trajectories} initial estimates of
body parts are retrieved. The trajectory for feature $i$ is the
spatial time series
$\mathbf{f}^0_i,\ldots,\mathbf{f}^{(N_\text{L}-1)\Delta T}_i$, where
$N_\text{L}$ is the number of frames and $\frac{1}{\Delta T}$ the
frame rate. $\mathcal{V}$ is the complete set of $N_\text{V}$
trajectories $\mathbf{v}_0,\ldots,\mathbf{v}_{N_\text{V}-1}$. The
pairwise trajectory distances are expressed via (modified
from~\citep{Krahnstoever03})
\begin{eqnarray}
d(\mathbf{v}_i,\mathbf{v}_j)&=&
\alpha\sum^{N_\text{L}-1}_{t=0}(\Delta_{ij}^t-\overline{\Delta}_{ij})^2
\notag\\ 
&+&(1-\alpha)\sum^{N_\text{L}-2}_{t=0}{\left(1-\left\langle\mathbf{vel}_i^{t},\mathbf{vel}_j^t\right\rangle\right)}
\label{eq:TrajectoryDistance}
\end{eqnarray}
with $\Delta_{ij}^t=\left\|\mathbf{f}_i^t-\mathbf{f}_j^t\right\|$,
$\mathbf{vel}^t_i=\mathbf{f}_i^{t+1}-\mathbf{f}_i^{t}$, and
$\overline{\Delta}_{ij}$ is the mean of $\Delta^t_{ij}$ over all
frames.  $\left\langle\cdot ,\cdot\right\rangle$ represents the scalar
product, $\alpha=0.01$.  

Eq.~(\ref{eq:TrajectoryDistance}) is plugged into a
\emph{self-tuning}~\citep{Manor04} framework to extract perceptually
consistent limb representations without manual
intervention. We employ iterative
\emph{normalized cut} clustering to segment the trajectory
dataset. Instead of
$\text{NCut}(\mathcal{V}_0,\mathcal{V}_1)$~\citep{Shi00} that splits the trajectory
set $\mathcal{V}$ in two child clusters we use
\begin{equation}
 \text{NCut}'(\mathcal{V}_0,\mathcal{V}_1)=\text{NCut}(\mathcal{V}_0,\mathcal{V}_1) e^{-\frac{\sigma_\text{b}^2-\alpha_{\text{b}}}{\beta_{\text{b}}}}\,,
\end{equation}
with empirically determined values of $\alpha_{\text{b}}=20.0$ and
$\beta_{\text{b}}=200.0$.  If this value exceeds
$\tau_{\text{NC}}=0.35$, or if the number of features in any child
cluster becomes $\leq 10$, splitting is stopped. With such a large
threshold, the trajectory sets may become oversegmented; excess
clusters are primarily caused by non-rigid cloth motion. Cloth-induced
clusters generally arise from distortion of attire, and rarely show
significant motion relative to the respective limbs. Clusters
representing true body parts tend to move vividly and to rotate
relative each other due to the rotatory connections enforced by the
underlying kinematic skeleton (cf.~\citep{Walther11}). As a
consequence, all feature clusters with a relative rotation of less
than $15^{\circ}$ and with a mutual barycenter shift of less than 10
pixels are merged, which reliably eliminates cloth-induced excess
clusters. Prior to the merging stage, statistical outlier removal
eliminates all features whose time course strays significantly
from that of their host cluster. The number of clusters after merging
is $N_{\text{G}}$, the clusters are identified by $\mathcal{G}_i$ and
shown for the reference frame in Fig.~\ref{fig:FeatureGroups}.

\subsection{Retrieving kinematic skeletons}
Recent skeleton extraction approaches
include~\citep{Pollefeys08,Ross10} -- for the current work, the one
by~\citep{Krahnstoever03} is favored. There, skeleton extraction is
applied to full limb templates, here we retrieve skeletons at an
earlier stage directly from feature group data; the quality of the
results is comparable to those of~\cite{Krahnstoever03}. We use a
different distance-based joint plausibility criterion:
\begin{equation}
s^k_{ij}=\min\limits_{u}\left\|\mathbf{x}_{ij,\text{wrl}}^{0*}-\mathbf{f}^0_u\right\|,\ 
\mathbf{f}^0_u\in\mathcal{G}_k,\ k\in\{i,j\}\,.
\label{eq:SpatialJointPlausibility}
\end{equation}
$\mathbf{x}_{ij,\text{wrl}}^{0*}$ is the world position at $t=0$ of a
putative joint between feature groups $\mathcal{G}i$ and
$\mathcal{G}j$.  Eq.~(\ref{eq:SpatialJointPlausibility}) causes
alteration of~\citeauthor{Krahnstoever03}'s (\citeyear{Krahnstoever03}) original values from $a_s^+=1$
and $a_s^-=10$ to $a_s^+=20$, $a_s^-=100$ in our implementation
(see~\citep{Walther11} for details). Fig.~\ref{fig:Skeleton} demonstrates the performance of this scheme on the
previously segmented input scenario
(Fig.~\ref{fig:FeatureGroups}).

\subsection{Generating limb templates}

Although sufficient for skeleton extraction, the sparse body part
patterns give only an approximation of true human limb shape and must
be fleshed to compact \emph{limb templates}. All pixels
$\mathbf{x}:P^{t^*}_\text{F}(\mathbf{x})=1$ of
$\mathbf{I}^{t^*}(\mathbf{x})$ (color frame) are assigned to limb
template $i$ using
\begin{align}	D_k(\mathbf{x}) &=\min\limits_{\mathbf{f}^{t^*}_j\in\mathcal{G}_k}\left\|\mathbf{f}^{t^*}_j-\mathbf{x}\right\|\notag\\
	i_{\text{for pixel} \mathbf{x}}&=\arg\min\limits_{i'=0,\ldots,N_\text{G}-1}D_{i'}(\mathbf{x})\,.
\end{align}
The resulting \emph{limb masks} are exemplarily depicted in
Fig.~\ref{fig:LimbSegmentation}, color templates for each body part
$i$ can easily be learned by collecting information in areas of
$\mathbf{I}^{t^*}(\mathbf{x})$ covered by the respective limb
masks. Shape templates are constructed by scanning the
outer perimeter of each mask, thus avoiding shape contributions from
the foreground area. The learned body part templates
do not take into account deformation behavior of human limbs, but
in downstream processing, such deformation will be averaged out
anyway.

\subsection{Pictorial structures for upper human body modeling}

It remains to cast the extracted templates and kinematic constraints
into a concise body pattern, which is represented by a pictorial
structure model~\citep{Fischler73,Huttenlocher00}.  Our upper-body PS
representations (Fig.~\ref{fig:SystemOverview}h) allow to unify
appearance and kinematic constraints of the observed subject. Each
model comprises a tree-like graph with vertices representing the
appearance of the body parts found in the limb extraction stage. Each
graph edge encodes the `bones' (kinematic constraints) of
the previously extracted skeleton. The PS is further augmented with an
array of joint angles learned for each retrieved body joint
incorporating sequence-specific limb orientation information.

A detailed discussion of pictorial structures and matching strategies
is left to~\citep{Huttenlocher00,Huttenlocher05,Walther11}. We
adopt matching techniques proposed
by~\citet{Huttenlocher00,Huttenlocher05} with the following
modifications: A `switched' slope parameter $k_{\theta}$ (which is
constant in~\citep{Huttenlocher00} allows for
integration of joint angle limits: if a joint's rotation angle is
within limits, $k_{\theta}$ remains small, ensuring nearly
unconstrained joint motion, otherwise $k_{\theta}$ is significantly
increased to penalize posture estimates violating learned
joint ranges. Additionally, \emph{kinematic flips}~\citep{Triggs03a}
are tolerated in PS matching, deviating from~\citep{Huttenlocher00}:
model limbs may flip around their principal axes to better accommodate
complex body postures. Joint angles are automatically adopted during
the flipping process, which is restricted to both forearms.

For registering the PS model the input color image
$\mathbf{I}(\mathbf{x})$ is first converted to a binary line image
using the \emph{EDISON} algorithm~\citep{Meer02}. Using \emph{oriented
  chamfer distance}~\citep{Shotton08a}, thinned (thinning
routines from~\citep{IPL06} representations of the learned limb
perimeters are matched to the EDISON-based line representation
(see~\citep{Walther11} for details of this procedure).

\section{Meta model formulation}
These PS models are highly scenario-specific and will be inadequate
for posture estimation in novel situations. A model learned from a
subject wearing a red short-sleeve shirt will likely fail to match
another subject wearing a green long-sleeve shirt.  Nevertheless, the
sequence-specific PS models created above might well be consolidated
to yield a more generic and powerful \emph{meta
  model}~\citep{Walther10} $\mathbf{M}_{\text{meta}}$ that represents
the upper human body on an abstract, conceptual level (Fig.~\ref{fig:SystemOverview}m).

To that end, let
$\mathcal{M}=[\mathbf{M}_0,\ldots,\mathbf{M}_{N_\text{M}-1}]$
represent an array of body models extracted from $N_\text{M}$
sequences using the techniques proposed above. For each model
$\mathbf{M}_i$
$\mathcal{S}_i=[\mathbf{s}_{i,0},\ldots,\mathbf{s}_{i,{N_\text{G}-1}}]$
is the array of smoothed and normalized shape templates. Smoothing is
of Gaussian type (using a standard deviation of 5 pixels) and helps to
cope with moderate cloth deformation, cf.~\citep{Walther11}. Subsequent
normalization forces all values in the smoothed template into $[0,1]$.
In addition, let
$\mathcal{C}_i=[\mathbf{c}_{i,0},\ldots,\mathbf{c}_{i,{N_\text{G}-1}}]$
constitute the color templates retrieved for each limb of
$\mathbf{M}_i$. Further, be $\mathcal{O}_{i,j}$ an array containing
all observed orientations of $\mathbf{M}_i$'s $j$\textsuperscript{th}
limb.  Any model $i$ contains a constant number $N_\text{J}$ of
joints, and comprises an array $\mathcal{W}_{i,k}$, which aggregates
all joint angles observed for joint $k$.

$\mathbf{M}_{\text{meta}}$ also holds a \emph{shape
  accumulator} array
$\mathcal{S}_{\text{acc}}=[\mathbf{s}_{\text{acc},0},\ldots,\mathbf{s}_{\text{acc},N_\text{G}-1}]$
and a \emph{color accumulator} array
$\mathcal{C}_{\text{acc}}=[\mathbf{c}_{\text{acc},0},\ldots,\mathbf{c}_{\text{acc},N_\text{G}-1}]$
for each \emph{meta limb}.  The accumulators for meta limb
$j$ are related to the meta model's body part representations
according to
\begin{equation}	\mathbf{s}_{\text{meta},j}=\frac{\mathbf{s}_{\text{acc},j}}{N_\text{I}},\  \mathbf{c}_{\text{meta},j}=\frac{\mathbf{c}_{\text{acc},j}}{N_\text{I}}\,.
\label{eq:PrototypeFormation}
\end{equation}
$N_\text{I}$ indicates the number of sequence-specific models already
integrated into $\mathbf{M}_{\text{meta}}$. There is also a
\emph{joint position accumulator} array, discussed in~\citep{Walther11}.

To initialize the meta model, let
$\mathbf{M}_{\text{meta}}=\mathbf{M}_0$. Limb templates from
$\mathbf{M}_0$ are copied into the accumulators of the meta model,
$N_{\text{I}}$ is accordingly set to 1. In addition, topology and
connectivity are cloned from $\mathbf{M}_0$, as well as joint limits
and angular distributions.

\subsection{Aligning the input models}
Integrating information from each $\mathbf{M}_i$ ($i>0$) into
the evolving meta model is a concept building task.
The meta limb prototypes (i.\,e., the limb templates of
$\mathbf{M}_{\text{meta}}$) are updated by sequentially adding
information from all $\mathbf{M}_i$ ($i>0$) to 
$\mathbf{M}_{\text{meta}}$. To ease that process, the structure of
each incoming model $\mathbf{M}_i$ is \emph{aligned} to match the
current meta model structure w.\,r.\,t.\ body part alignment, limb
enumeration, and joint enumeration. To that end, $\mathbf{M}_i$ is
instantiated to take on a \emph{typical posture} by setting each of
the model's joints to the \emph{center angle} halfway in between the
links' upper and lower limits. \emph{Directional
  statistics}~\citep{Mardia00} are used to find these center
angles. The designated root limb acts as an anchor in the
instantiation process and is fixed to its mean orientation. Thinned
limb shapes of the instantiated model are then projected to an
artificial query image $I_\text{a}(\mathbf{x})$.  Barycenter
coordinates of each projected shape template $\mathbf{s}_{i,j}$ are
stored in $\mathbf{b}_{i,j}$. The current $\mathbf{M}_{\text{meta}}$
is matched to $I_\text{a}(\mathbf{x})$, using the PS
matching routines discussed in~\citep{Walther11}. After matching,
barycenters $\mathbf{b}_{\text{meta},k}$ of the meta model should
project near the $\mathbf{b}_{i,j}$ if only if meta limb $k$ corresponds to
limb $j$ in $\mathbf{M}_i$. Then meta limb $k$ is defined to
correspond to model limb
\begin{equation}	j=\arg\min\limits_{j'}\left\|\mathbf{b}_{\text{meta},k}-\mathbf{b}_{i,j'}\right\| \label{eq:CorrespondenceProblem}
\end{equation}

\noindent Knowing all limb correspondences, $\mathbf{M}_{i}$ can
readily be manipulated to comply with the meta model's current
structure: limb and joint enumeration are unified by reindexing, using
the retrieved correspondences. After reindexing, limb $j$ in model $i$
corresponds to meta limb $j$ and joint $k$ of $\mathbf{M}_i$
corresponds to meta joint $k$. With that, body-centric coordinate
systems of all limbs in $\mathbf{M}_i$ are adjusted such that limb
focusing in $\mathbf{M}_i$ and $\mathbf{M}_\text{meta}$ becomes
identical; thorough bookkeeping is necessary to keep values in
each $\mathcal{O}_{i,j}$ and each $\mathcal{W}_{i,k}$ consistent.
After these preparations, $\mathbf{M}_i$ is 
\emph{aligned} with the current meta model.

\subsection{Learning meta shape prototypes}

Formulating prototypical shapes (Fig.~\ref{fig:SystemOverview}i) for a
structure that deforms as vividly as a dressed human limb is not
trivial, we rely on the approximate registration between each shape
template $\mathbf{s}_{i,j}$ of $\mathbf{M}_i$ and the corresponding
meta shape prototype $\mathbf{s}_{\text{meta},j}$.  Based on that, a
\emph{registration operator} applies the 2D \emph{iterative closest
  point} (ICP) method of~\cite{Besl92}, accelerated according
to~\cite{Levoy01}, to compensate for the residual alignment failure
between the shape representations. Following ICP registration,
$\mathbf{s}_{i,j}$ and $\mathbf{s}_{\text{meta},j}$ are assumed to be
aligned optimally in the sense of~\cite{Besl92}.  The aligned
$\mathbf{s}_{i,j}$ is eventually summed into
$\mathbf{s}_{\text{acc},j}$; this summation for all $i>0$
yields a \emph{voting process}~\citep{Grauman09}: shape
pixels strongly voted for by constantly high accumulator values evolve
into \emph{persistent outline modes}, whereas areas not supported by
given evidence fade out during aggregation. After adding
$\mathbf{s}_{N_\text{M}-1,j}$, the weakest 25 percent of the collected
votes are removed from the accumulator in order to memorize only
reliable outline segments for each processed body part.

\subsection{Acquiring meta color prototypes}

Learning meta color prototypes (Fig.~\ref{fig:SystemOverview}j) is
more involved than shape prototype construction: assuming that color
information from some sequence-specific model $\mathbf{M}_i$ shall be
exploited to update the meta color prototypes, results from above
matching/ICP registration can be carried forward to align color
prototype $\mathbf{c}_{i,j}$ with the corresponding color accumulator
$\mathbf{c}_{\text{acc},j}$. The aligned color representations are
mapped to HSV color space  and pixelwise color similarities
are measured by HS-histogram-based windowed correlation. The
V-component is dropped in order to increase robustness against
illumination variation~\citep{Hu09}.  A binary \emph{persistent color
  mask} $M_{i,j}(\mathbf{x})$ is then defined such that pixels in
$M_{i,j}(\mathbf{x})$ take on `1' values if and only if the
correlation result at image location $\mathbf{x}$ exceeds a threshold
of 0.25.  Guided by $M_{i,j}(\mathbf{x})$, information from
$\mathbf{c}_{i,j}$ is used to update the meta model's
$j$\textsuperscript{th} color accumulator, according to
\begin{equation}
\mathbf{c}_{\text{acc},j}(\mathbf{x})=
\begin{cases}
    \mathbf{c}_{\text{acc},j}(\mathbf{x})+\mathbf{c}_{i,j}(\mathbf{x})  &
    \text{if } M_{i,j}(\mathbf{x})>0\\ 
    \mathbf{0} & \text{otherwise}
\end{cases} \,.
\label{eq:ColorAggregation}
\end{equation}
%
When applied to all $\mathbf{c}_{i>0,j}$,
eq.~\ref{eq:ColorAggregation} suppresses color information that varies
significantly between sequences. Persistent colors are preserved as
desired and yield, via eq.~(\ref{eq:PrototypeFormation}), the prototypes
$\mathbf{c}_{\text{meta},j}, j \in\{0,\ldots,N_{\text{G}-1}\}$. Any $\mathbf{c}_{\text{meta},j}$ is
considered \emph{valid} if it contains at least one nonzero pixel.

\begin{figure*}[!h]
{\Large\bf Additional Material}\\[5mm]

\centering
\subfloat[Original image \label{fig:BackprojectionImage} \label{fig:Backprojection_OriginalImage}]{\includegraphics[width=0.45\textwidth]{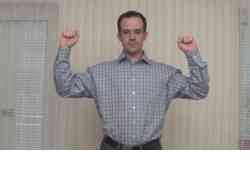}}
	\hspace{0.15cm}
\subfloat[Results of backprojecting information from $\mathcal{H}_{\text{meta},\text{torso}}$ to Fig.~\ref{fig:Backprojection_OriginalImage}. \label{fig:Backprojection_ResultImage} ]{\includegraphics[width=0.45\textwidth]{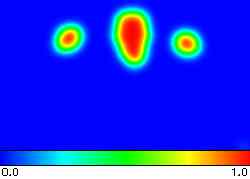}}
	\caption[Exemplary backprojection map]{Exemplary backprojection map, color encoding is in \emph{heat map} style, as indicated by the color reference bar below Fig.~\ref{fig:Backprojection_ResultImage}.}
\end{figure*}

Now each $\mathbf{c}_{\text{meta},j}$ is augmented with an
HS-histogram $\mathcal{H}_{\text{meta},j}$ that allows for
\emph{histogram backprojection}~\citep{Ballard91}. To populate
$\mathcal{H}_{\text{meta},j}$, a complex sampling scheme is employed;
see~\citep{Walther11} for details.  Backprojecting
$\mathcal{H}_{\text{meta},j}$ to novel image content yields the
\emph{backprojection map} $C_j(\mathbf{x})$ for the corresponding body
part. Windowed histogram backprojection is employed here in order to
increase compactness of the generated maps.  For posture estimation
the backprojection maps are thresholded at 10\% of their peak and
blurred by a Gaussian with a standard deviation of 5.0 pixels.
Follow-up normalization forces $C_j(\mathbf{x})$ into $[0,1]$, all
entries in backprojection maps corresponding to meta limbs without
valid color prototype are set to 1.0.
Fig.~\ref{fig:Backprojection_ResultImage} (additional material)
exemplarily shows $C_\text{torso}(\mathbf{x})$ resulting from
backprojection of $\mathcal{H}_{\text{meta},\text{torso}}$ to the
query image in Fig.~\ref{fig:Backprojection_OriginalImage} (additional
material).  Based on $C_j(\mathbf{x})$, the \emph{color cue map}
$C_{\theta_j,s_j}(\mathbf{x})$ for meta limb $j$ with orientation
$\theta_j$ and scale $s_j$ is readily defined: let
$\mathbf{c}'_{\text{meta},j}$ be a binarized representation of
$\mathbf{c}_{\text{meta},j}$, with
$\mathbf{c}'_{\text{meta},j}(\mathbf{x})=1$ if and only if
$\left\|\mathbf{c}_{\text{meta},j}(\mathbf{x})\right\|>0$. Then be
$\mathbf{c}'_{\text{meta},\theta_j,s_j}$ an instance of
$\mathbf{c}'_{\text{meta},j}$, oriented and scaled as to match meta
limb $j$'s desired state. With that
\begin{equation}	
C_{\theta_j,s_j}(\mathbf{x})=
\frac{C_j(\mathbf{x})*\mathbf{c}'_{\text{meta},\theta_j,s_j}}{\sum\mathbf{c}'_{\text{meta},\theta_j,s_j}}\,,
\end{equation} 
where `$*$' is convolution and the sum aggregates all nonzero pixels
in $\mathbf{c}'_{\text{meta},\theta_j,s_j}$.

\subsection{Gabor prototype generation}
Besides shape and color information \emph{persistent texture} can be
learned autonomously from given input data (Fig.~\ref{fig:SystemOverview}k). To that end, we employ
\emph{Gabor wavelets} as tuneable, localized frequency filters in the
construction of \emph{Gabor grid graphs} for each meta limb $i$.
Graph nodes correspond to mean Gabor magnitude \emph{jets}
(cf.~\citep{transcomp})
$\mathbf{J}_{i,j},\ j=0,\ldots,N_{\text{Q},i}-1$ learned from the
input streams. The jet learning scheme uses the same batch process as
employed for color prototyping (cf.~\citep{Walther11}). Given the mean
jets, batch learning is restarted to calculate each jet $j$'s
\emph{covariance matrix}, whose largest eigenvalue $\lambda^*_{i,j}$
provides a convenient measure of the node's
\emph{reliability} $\eta_{i,j}=1/\sqrt{\lambda^*_{i,j}}$. In our
approach, two normalized complex jets $\mathbf{J}_\text{A}$ and
$\mathbf{J}_\text{B}$ are compared by inspecting their absolute parts
only~\citep{transcomp}.
\begin{equation}
S_{\text{Abs}}\left(\mathbf{J}_\text{A},\mathbf{J}_\text{B}\right)
=\sum_j a_{\text{A},j} a_{\text{B},j} 
\label{eq:AbsSimFunction}
\end{equation}
Being exclusively interested in persistent texture, all nodes with
reliabilities $\eta_{i,j}<5.0$ are pruned. The largest connected
subgraph $G^*_{\text{G},i}$ that survives this procedure makes
up a valid \emph{Gabor prototype} for meta limb $i$ if and only if its node count
$N^*_{\text{Q},i}$ exceeds 50; this large threshold restricts
prototype learning to \emph{meaningful} Gabor patches. Henceforth,
define the nodes of prototype graph $i$ to be
$\mathbf{g}_{i,j},\ j=0,\ldots,N^*_{\text{Q},i}-1$. In our
experiments, a valid Gabor prototype evolved exclusively on the head
region; potential Gabor patterns for all other limbs were depleted of
nodes by pruning and became invalid. Note that our system
learns to treat the head and the thorax region of an observed subject
as a single \emph{torso} entity, because the training data did not
include movement between the two. Thus, the evolved Gabor prototype
can be seen as a generic \emph{texture-based torso detector} that
optimally responds to human torsi in upright position.

\begin{figure*}[htb]
	\centering	\subfloat[Original image\label{fig:TorsoDetectionA}]{\includegraphics[width=0.45\textwidth]{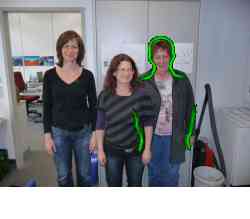}}
	\hspace{0.15cm}	\subfloat[Gabor cue map for Fig.~\ref{fig:TorsoDetectionA} \label{fig:TorsoDetectionB}]{\includegraphics[width=0.45\textwidth]{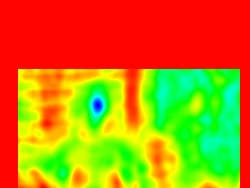}}
	\caption[Gabor cue map]{Exemplary Gabor cue map for the meta
          torso: for visual convenience, normalization to $[0;,1]$ has
          been applied in the right image; generic heat map color
          encoding is used, black regions represent `forbidden' torso
          locations. The torso detection corresponding to the most
          pronounced minimum in Fig.~\ref{fig:TorsoDetectionB} has
          been overlaid to Fig.~\ref{fig:TorsoDetectionA} (green
          outline).} \label{fig:TorsoDetector}
\end{figure*}

A \emph{Gabor jet representation}
$\mathcal{G}_I\left(\mathbf{x}\right)$ of query image $I(\mathbf{x})$
yields a \emph{Gabor cue map} for meta limb $i$ in orientation
$\theta_i$ and with scale~$s_i$:
\begin{eqnarray}	
G_{\theta_i,s_i}(\mathbf{x})&=&
1-\frac{1}{N^*_{\text{Q},i}}\sum\limits_{\mathbf{g}_{i,j}\in G^*_{\text{G},i}}S_\text{Abs}\left(\mathbf{J}_{i,j},\mathbf{J}_I\right),\notag\\
 \mathbf{J}_I&=&\mathcal{G}_I\left(\mathbf{x}+\mathcal{P}_{\theta_i,s_i}\left(\mathbf{g}_{i,j}\right)\right)\,,
\end{eqnarray}
where $\mathcal{P}_{\theta_i,s_i}\left(\cdot\right)$ projects nodes of
$G^*_{\text{G},i}$ into $I(\mathbf{x})$.
Fig.~\ref{fig:TorsoDetector} (additional material) demonstrates application of the learned
torso detector (in upright position and with scale 1.0) to the image
in Fig.~\ref{fig:TorsoDetectionA}: the observed torso barycenters
correspond to pronounced minima in the Gabor cue map depicted
in Fig.~\ref{fig:TorsoDetectionB}.

\subsection{Limb pre-detection}
Rotation of any meta limb $i$ with valid Gabor prototype
$G^*_{\text{G},i}$ can safely be assumed
negligible~\citep{Walther11}. This allows to condense the orientation
dimension of the meta limb's state space to a single, \emph{typical}
value $\theta_{\text{typ},i}$ which corresponds to the mean of all
body part orientations observed during model learning.  Accordingly,
we define the \emph{color-augmented Gabor detection map} for meta limb
$i$, presuming that a valid $G^*_{\text{G},i}$ exists
\begin{equation}
\overline{G}_{\theta_{\text{typ},i},s_i}(\mathbf{x})=
  \begin{cases}
\frac{1}{2}G(\mathbf{x}) & 
\text{if } G(\mathbf{x})<\frac{\widehat{G}}{2}
\text{ and } C(\mathbf{x})>\frac{\widehat{C}}{2}\\
G(\mathbf{x}) & \text{otherwise}\,,
\end{cases}
 \label{eq:GaborLimbDetection}
\end{equation}
where $G(\mathbf{x})$ is a shortcut for
$G_{\theta_{\text{typ},i},s_i}(\mathbf{x})$, $C(\mathbf{x})$ stands
for $C_{\theta_{\text{typ},i},s_i}(\mathbf{x})$; $\widehat{G}$ is the
peak value of $G_{\theta_{\text{typ},i},s_i}(\mathbf{x})$ and
$\widehat{C}$ represents the largest value of
$C_{\theta_{\text{typ},i},s_i}(\mathbf{x})$.  Combining the two `weak'
cues in eq.~(\ref{eq:GaborLimbDetection}), the system successfully
eliminates wrong detection optima in all performed experiments. With
that,
\begin{equation}	
s_{\text{pre},i}=\arg\min\limits_{s}\left(\min\limits_{\mathbf{x}}\overline{G}_{\theta_{\text{typ},i},s}(\mathbf{x})\right) 
\label{eq:ScaleFinding}
\end{equation}
yields the most probable scale estimate for meta limb $i$ given
texture and color evidence in input image $\mathbf{I}(\mathbf{x})$.
Minimization in eq.~(\ref{eq:ScaleFinding}) is performed with 30 discrete
scales between 0.7 and 1.0. Going beyond scale,
let
\begin{equation}
\mathbf{x}_{\text{pre},i}=\arg\min\limits_{\mathbf{x}}\overline{G}_{\theta_{\text{typ},i},s_{\text{pre},i}}(\mathbf{x})
\end{equation}
and eventually assume meta limb $i$ to be \emph{pre-detected} in
location
$\mathbf{l}_{\text{pre},i}=\left(\mathbf{x}_{\text{pre},i},\theta_{\text{typ},i},s_{\text{pre},i}\right)$.
In the following experiments, positional search space for any
pre-detectable meta limb $i$ is restricted to $\pm 10$ pixels around
$\mathbf{x}_{\text{pre},i}$. Such body part \emph{anchoring} acts, via
given articulation constraints, on the complete meta model and allows
to circumnavigate false local posture optima induced by background
clutter. Thus, limb anchoring renders the final posture estimate more
robust. Higher processing speed can be expected as a positive side
effect of the anchoring procedure, as fewer state space locations have
to be probed in the model matching cycle.

\subsection{Enforcing color constancy}
Scenarios with unrestricted illumination conditions require \emph{chromatic
  adaptation}~\citep{Suesstrunk00} like human
perception~\citep{Paris08} to achieve approximate \emph{color
  constancy} (Fig.~\ref{fig:SystemOverview}l).  This relies on persistently colored parts of the body to act as
\emph{intrinsic color reference} for autonomous chromatic
adaptation, similar to~\citep{Montojo09}: first, input
image $\mathbf{I}(\mathbf{x})$ is transformed to \emph{Lab opponent
  color space}\citep{Margulis06,Montojo09}, yielding
$\mathbf{I}_{\text{Lab}}(\mathbf{x})$.  Lab space allows to remove unwanted color
deviations by balancing the each pixel's \emph{chromaticity
  coordinates} until the cast
vanishes while leaving luminance values largely unaltered.
A \emph{shift vector} of length $R_i\in\{0,2,4,8,16,32\}$ and direction 
$\gamma_i\in\{v_i\frac{2\pi}{8}:v_i=0,\ldots,7\}$ is added to
the (a,b)-components of each pixel in
$\mathbf{I}_{\text{Lab}}(\mathbf{x})$ and results in a
\emph{color-shifted} Lab input image
$\mathbf{I}_{\text{Lab},R_i,\gamma_i}(\mathbf{x})$. Conversion of
$\mathbf{I}_{\text{Lab},R_i,\gamma_i}(\mathbf{x})$ to HSV yields
$\mathbf{I}_{\text{HSV},R_i,\gamma_i}(\mathbf{x})$. A windowed
backprojection (window size: 7x7 pixels) of
$\mathcal{H}_{\text{meta},i}$ onto
$\mathbf{I}_{\text{HSV},R_i,\gamma_i}(\mathbf{x})$ gives the
\emph{color similarity map} $U_{R_i,\gamma_i}(\mathbf{x})$, whose
values are normalized to $[0;1]$.

\begin{figure*}[tb]
	\centering
	\subfloat[Low entropy]{\includegraphics[width=0.45\textwidth]{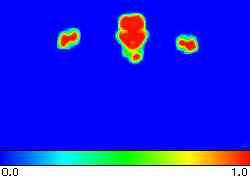}}
	\hspace{0.15cm}
	\subfloat[High entropy]{\includegraphics[width=0.45\textwidth]{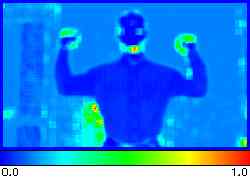}}
	\caption[Entropy states]{Visualized entropy states: the depicted color similarity maps $U_{R_\text{torso},\gamma_\text{torso}}(\mathbf{x})$ are calculated for the original input image in Fig.~\ref{fig:Backprojection_OriginalImage}\label{fig:EntropyResults}; color encoding is in heat map style.}
\end{figure*}

Assuming existence of a valid Gabor prototype (and thus a valid
$\mathbf{l}_{\text{pre},i}$) for meta limb $i$, the binary
$\mathbf{c}'_{\text{meta},i}$ are projected into the image plane (according to the
parameters in $\mathbf{l}_{\text{pre},i}$); morphological
opening follows to get rid of noise in the induced \emph{projection
  map} $\mathbf{c}'_{\text{meta},\mathbf{l}_{\text{pre},i}}$.
From that we define the \emph{color similarity measure}
\begin{equation}
	C_{R_i,\gamma_i}=\frac{\sum\limits_{\mathbf{x}:h(\mathbf{x})=1}
          U_{R_i,\gamma_i}(\mathbf{x})}{\sum\mathbf{c}'_{\text{meta},\mathbf{l}_{\text{pre},i}}} \label{eq:CAColorSimilarityMeasure}\,,
\end{equation}
where $h(\mathbf{x})=\mathbf{c}'_{\text{meta},\mathbf{l}_{\text{pre},i}}(\mathbf{x})$.
Additionally, approximate a probability distribution
\begin{equation}
p_{R_i,\gamma_i}(\mathbf{x})=\frac{U_{R_i,\gamma_i}(\mathbf{x})}{\sum{U_{R_i,\gamma_i}}}
\end{equation}
with $\sum{U_{R_i,\gamma_i}}$ being the sum of all entries in the color similarity map.
This allows to set up the \emph{entropy}
of the color similarity map, according to 
\begin{equation}
S_{R_i,\gamma_i}=-\sum\limits_{\mathbf{x}\in\mathcal{X}} p_{R_i,\gamma_i}(\mathbf{x})\ln(p_{R_i,\gamma_i}(\mathbf{x}))\,.
\end{equation}
$S_{R_i,\gamma_i}$ grows large if the color distribution in
$U_{R_i,\gamma_i}(\mathbf{x})$ becomes diffuse, well-defined
clusters bring the entropy down (Fig.~\ref{fig:EntropyResults}, additional material). As persistent color
patches stored in the meta limbs generally constitute coherent,
compact structures, their footprints in $U_{R_i,\gamma_i}(\mathbf{x})$
should (in the case of correctly chosen hyperparameters $R_i,\gamma_i$)
become blob-like; thus, color distributions with lower entropy 
are preferable.  Optimal \modified{hyperparameters}
$(R^*_i,\gamma^*_i)$ are found according to
\begin{equation}
(R^*_i,\gamma^*_i)=\arg\max\limits_{R_i,\gamma_i} \frac{C_{R_i,\gamma_i}}{S_{R_i,\gamma_i}}\,. \label{eq:CAParameterFinding}
\end{equation}
The corresponding Lab space shift vector that causes image colors
(after conversion to RGB) to near persistent colors of meta limb $i$
to the utmost possible extent (w.\,r.\,t.\ color similarity and
entropy) is defined as
$\mathbf{s}^*_i=\mathbf{s}_{R_i^{*},\gamma_i^{*}}$.
Fig.~\ref{fig:ColorCorrection} (additional material) shows the
efficiency of the chromatic adaptation routines employed here: the
depicted backprojection maps for the torso's histogram
$\mathcal{H}_{\text{meta},\text{torso}}$ show significant improvement
due to automatic color correction.

\subsection{Augmenting the matching cost function}
\begin{figure*}[t]
	\centering
	\subfloat[Without stimulus maps\label{fig:StimulusMapsA}]{\includegraphics[width=0.45\textwidth]{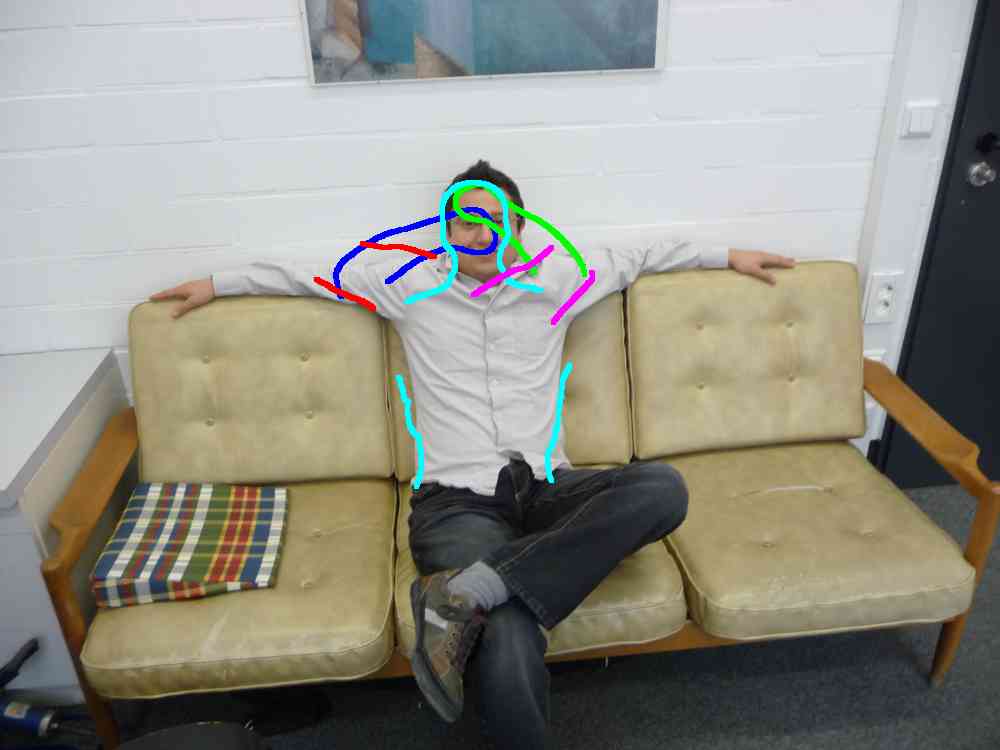}}
	\hspace{0.1cm}	\subfloat[Stimulus maps applied\label{fig:StimulusMapsB}]{\includegraphics[width=0.45\textwidth]{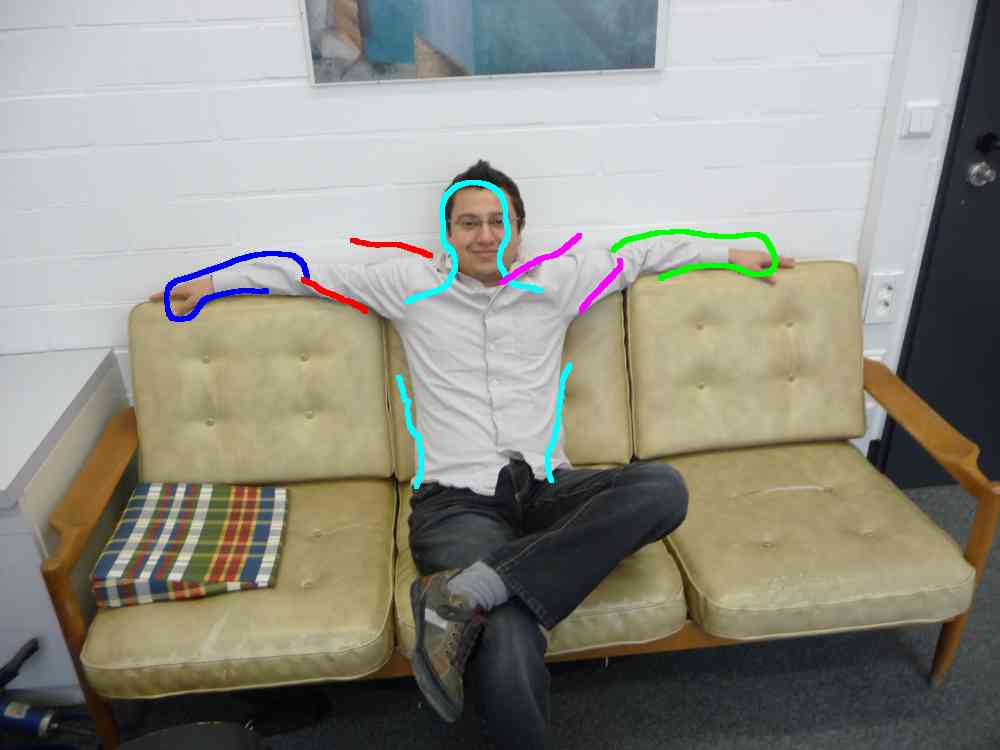}}\\
	\caption[Application of stimulus maps]{Without stimulus maps,
          double-counting~\citep{Zisserman08a} can yield
          wrong posture estimates
          (s. Fig.~\ref{fig:StimulusMapsA}). Activating the stimulus
          maps (s. Fig.~\ref{fig:StimulusMapsB}) corrects this issue
          and yields perceptually valid
          results. \label{fig:StimulusMaps}}
\end{figure*}

Given chromatically corrected input material, color information can be
used to enhance the PS model matching procedure so far based on pure
shape information. We define a \emph{negative stimulus map}
$N_i(\mathbf{x})$ that encodes a standard distance transformation on a
morphologically opened, inverted instance of
$\mathbf{c}'_{\text{meta},\mathbf{l}_{\text{pre},i}}$.  Subsequent
normalization forces elements of $N_i(\mathbf{x})$ to lie in [0;1].
The complementary \emph{positive stimulus map} is defined by
$P_i(\mathbf{x})=1-N_i(\mathbf{x})$. The negative stimulus map is
truncated at 0.3 to limit its influence on distant image structures.
With that, a range of \emph{spatially biased} backprojection maps for
all meta limbs is initialized as
$\overline{C}_i(\mathbf{x})=C_i(\mathbf{x})$, with
$i=0,\ldots,N_\text{G}-1$. For any pre-detectable meta limb $i$, these
maps are updated according to
\begin{eqnarray}
\overline{C}_j(\mathbf{x}) &=&
  \begin{cases}
    \overline{C}_j(\mathbf{x})\cdot P_j(\mathbf{x}) & \text{if } j=i\\
    \overline{C}_j(\mathbf{x})\cdot N_j(\mathbf{x}) & \text{if } j\neq i
  \end{cases}\,,
\label{eq:ColorManipulation2}
\\ \overline{C}_{\theta_i,s_i}(\mathbf{x}) &=&
\frac{\overline{C}_i(\mathbf{x})*\mathbf{c}'_{\text{meta},\theta_i,s_i}}{\sum\mathbf{c}'_{\text{meta},\theta_i,s_i}}
\end{eqnarray}
These modified color cue maps give rise to a
color-augmented matching cost function (see~\citep{Walther11} for
details)
\begin{align}
m_i(\mathbf{l}_i,\mathbf{I}(\mathbf{x}))=-\log\big(\big.&\left[1-S_{\theta_i,s_i}(\mathbf{x})\right]\cdot\notag\\
&\left.\left(0.65\cdot\overline{C}_{\theta_i,s_i}(\mathbf{x})+0.35\right)\right)\,,
\label{eq:RevisitedObservationScore}
\end{align}
where scaling and offsetting prevent color from dominating shape
information.  As shown in Fig.~\ref{fig:StimulusMaps}, this can
disambiguate complex \emph{double-counting}~\citep{Zisserman08a}
situations.

\begin{table}[t]
\centering
	\begin{tabular}{|c||r|r||r|r|}
	\hline
Benchmark &\multicolumn{2}{c||}{Permutation}&\multicolumn{2}{c||}{Redundancy}\\
		image & $\mu_{\text{P},b}$ & $\sigma_{\text{P},b}$ & $\mu_{\text{R},b}$ & $\sigma_{\text{R},b}$ \\
		\hline                                                                       
		A & 9.8 & 0.92 &    9.3 & 0.3 \\    
		\hline                     
		B & 9.1 & 3.8  &     6.6 & 0.5 \\    
		\hline                     
		C & 8.8 & 1.8  &     8.9 & 0.5 \\    
		\hline                     
		D & 11.2 & 1.4  &     10.8 & 0.4 \\    
		\hline                     
		E & 9.5 & 2.3  &     9.5 & 0.3 \\    
		\hline                     
		F & 7.1 & 1.4  &      6.5 & 0.5 \\    
		\hline
	\end{tabular}
\caption[Permuting meta model assembly order]{Mean and standard deviation of matching
results when permuting meta model
  assembly order (cols.~2, 3) and influence of redundant
  information (cols.~4, 5) on selected images (col.~1)}
\label{tbl:PermutationAndRepeatedLoadResults}
\end{table}

\section{Experimental evaluation}
\label{section:4}

\begin{figure*}[p]
	\centering	
\subfloat[\label{fig:INIPUREResultsA_imgA}]{\includegraphics[width=0.45\textwidth]{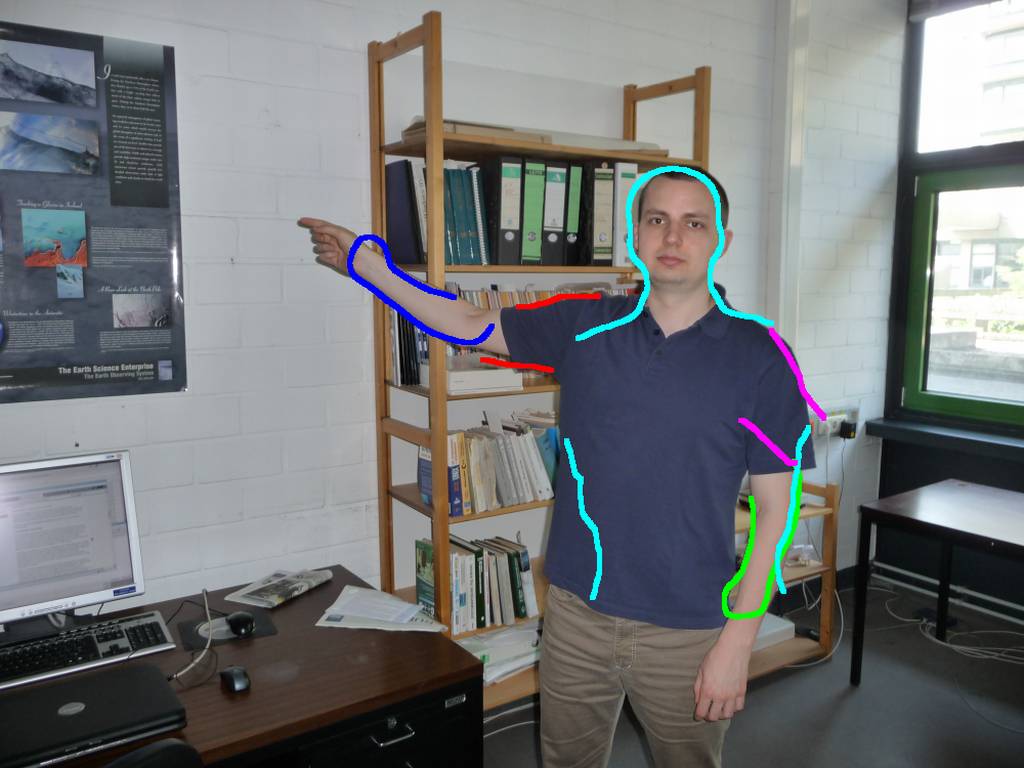}}
\hspace{0.1cm}
\subfloat[\label{fig:INIPUREResultsA_imgB}]{\includegraphics[width=0.45\textwidth]{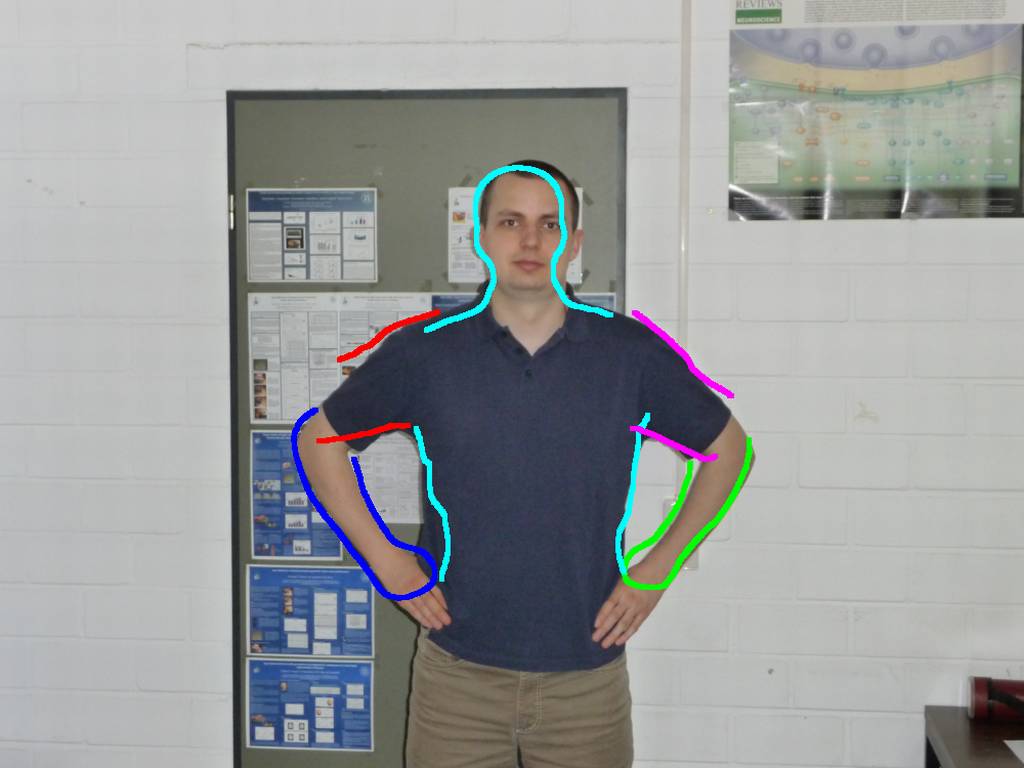}}\\

\subfloat[\label{fig:INIPUREResultsA_imgC}]{\includegraphics[width=0.45\textwidth]{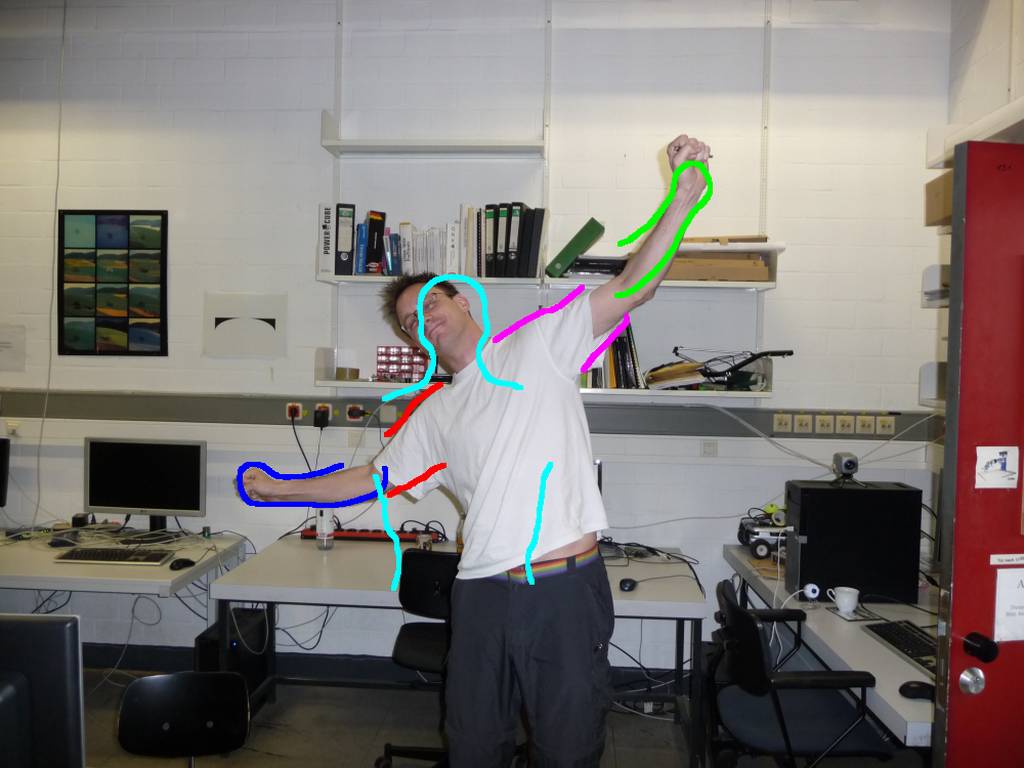}}
\hspace{0.1cm}
\subfloat[\label{fig:INIPUREResultsA_imgD}]{\includegraphics[width=0.45\textwidth]{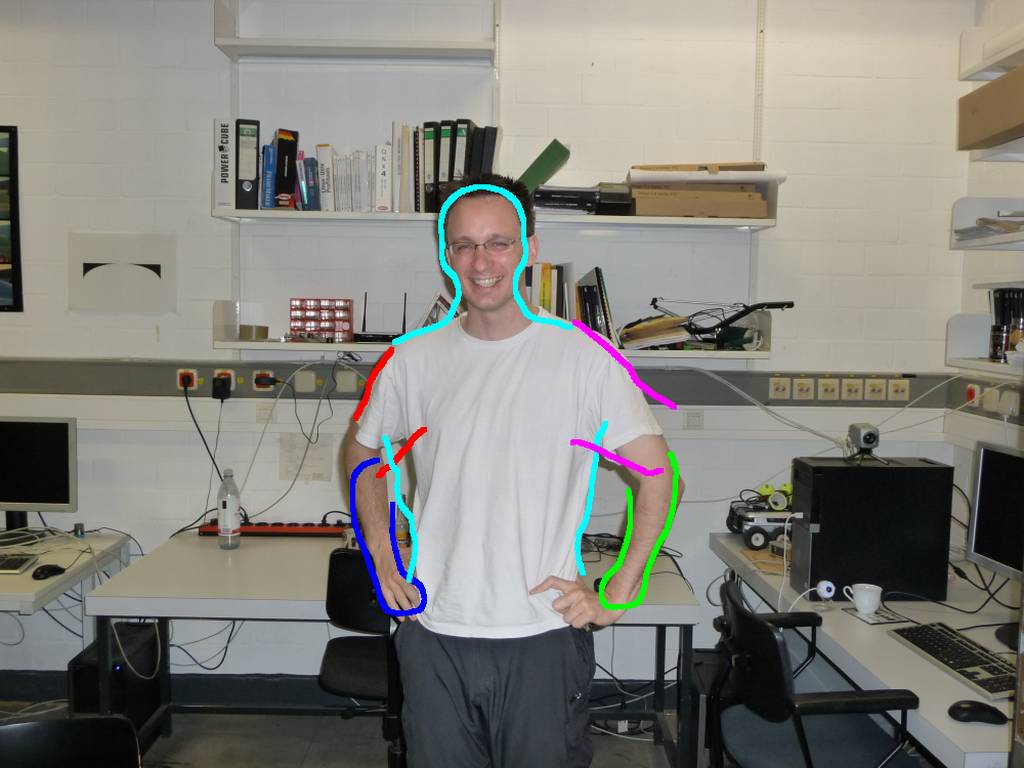}}\\

\subfloat[\label{fig:INIPUREResultsA_imgE}]{\includegraphics[width=0.45\textwidth]{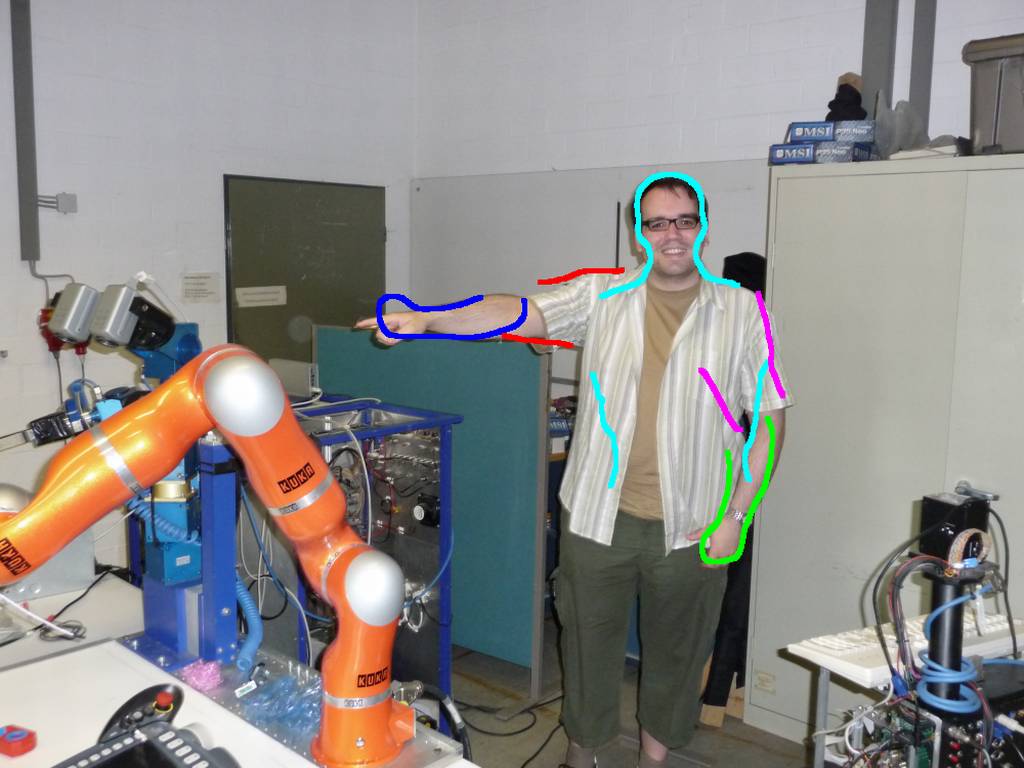}}
\hspace{0.1cm}
\subfloat[\label{fig:INIPUREResultsA_imgF}]{\includegraphics[width=0.45\textwidth]{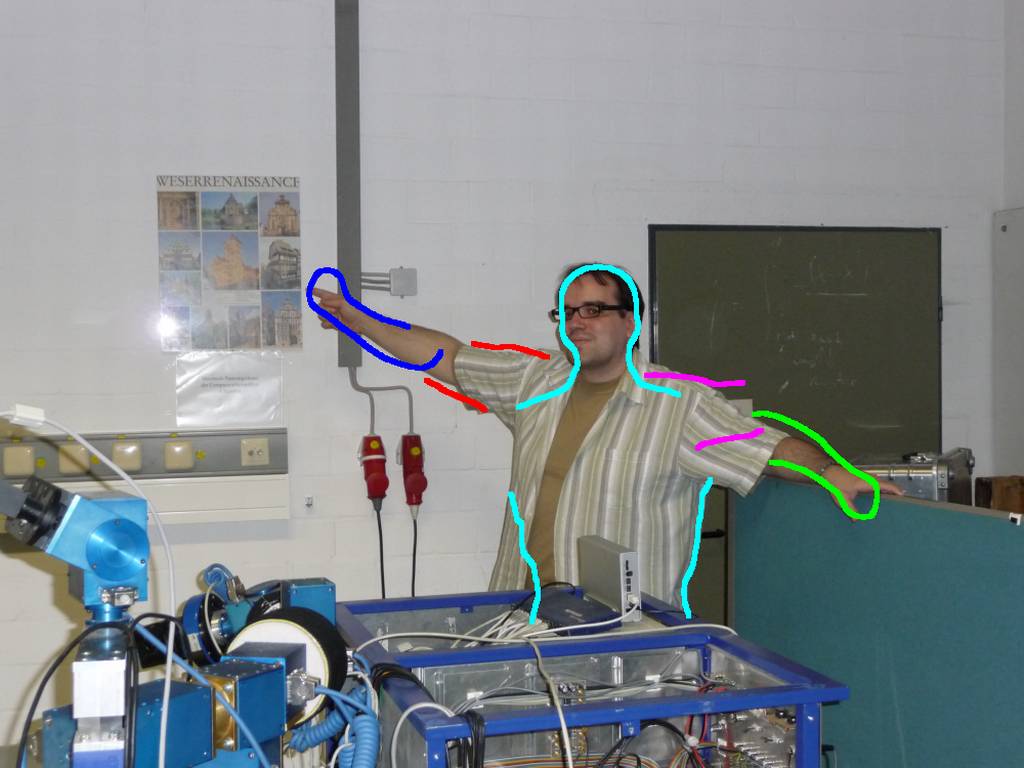}}\\

	\caption[Posture estimation results (sheet A)]{Posture estimation results (sheet A)}
	\label{fig:INIPUREResultsPlateA}
\end{figure*}

\begin{figure*}[tb]
	\centering
	\subfloat[Original input image\label{fig:CARealInput021}]{\includegraphics[width=0.45\textwidth]{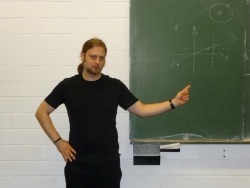}}
	\hspace{0.1cm}
	\subfloat[$C_\text{torso}(\mathbf{x})$ for Fig.~\ref{fig:CARealInput021} \label{fig:CARealColormapPreCA021}]{\includegraphics[width=0.45\textwidth]{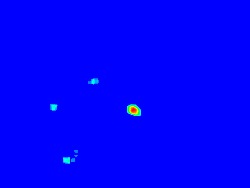}}\\
	\subfloat[After chromatic adaptation\label{fig:CARealOutput021}]{\includegraphics[width=0.45\textwidth]{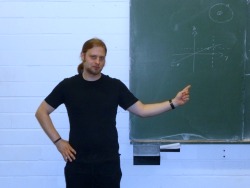}}
	\hspace{0.1cm}
	\subfloat[$C_\text{torso}(\mathbf{x})$ for Fig.~\ref{fig:CARealOutput021} \label{fig:CARealColormapPostCA021}]{\includegraphics[width=0.45\textwidth]{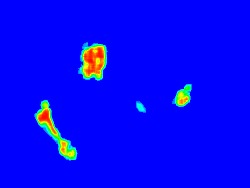}}\\
	\caption[Chromatic adaptation]{Exemplary chromatic adaptation results: the backprojection maps' color encoding is in heat map style; observe the significant improvement achieved through chromatic adaptation.\label{fig:ColorCorrection}}
\end{figure*}

\begin{figure*}[p]
	\centering	
\subfloat[\label{fig:INIPUREResultsK_imgA}]{\includegraphics[width=0.45\textwidth]{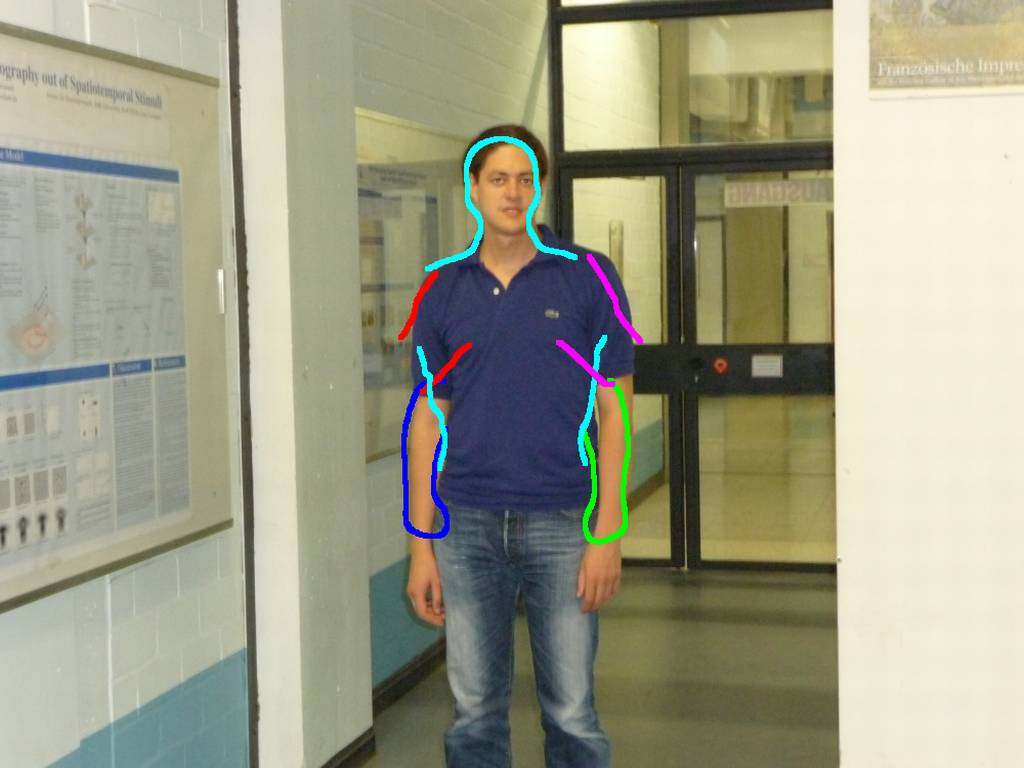}}
\hspace{0.1cm}
\subfloat[\label{fig:INIPUREResultsK_imgB}]{\includegraphics[width=0.45\textwidth]{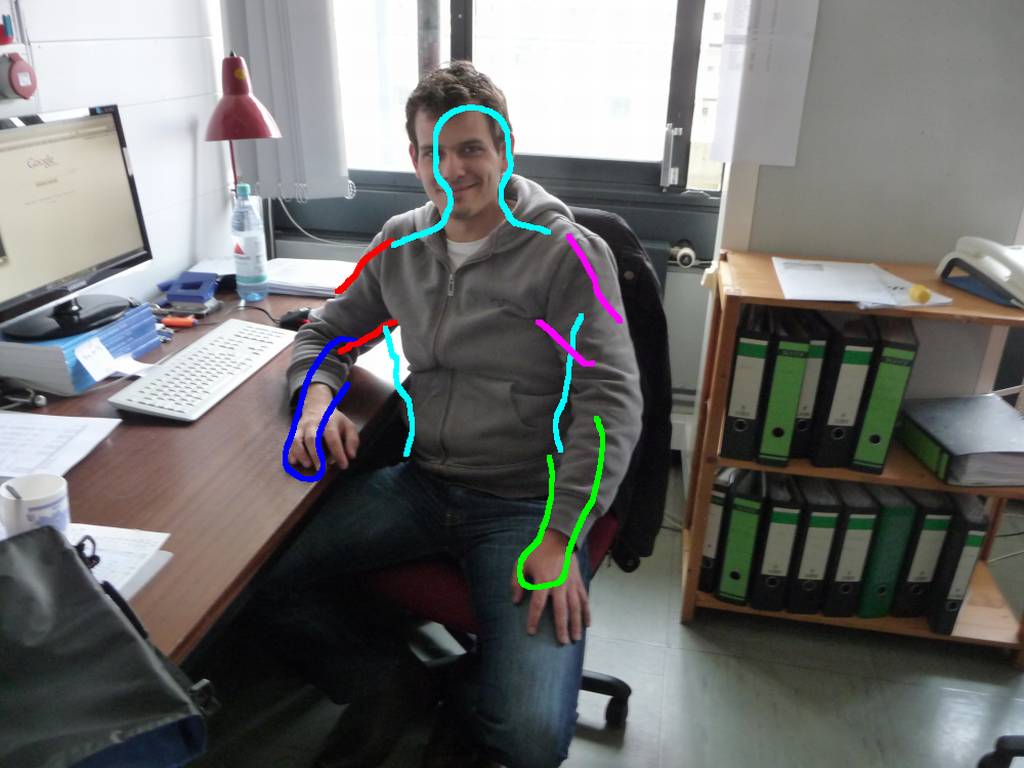}}\\

\subfloat[\label{fig:INIPUREResultsK_imgC}]{\includegraphics[width=0.45\textwidth]{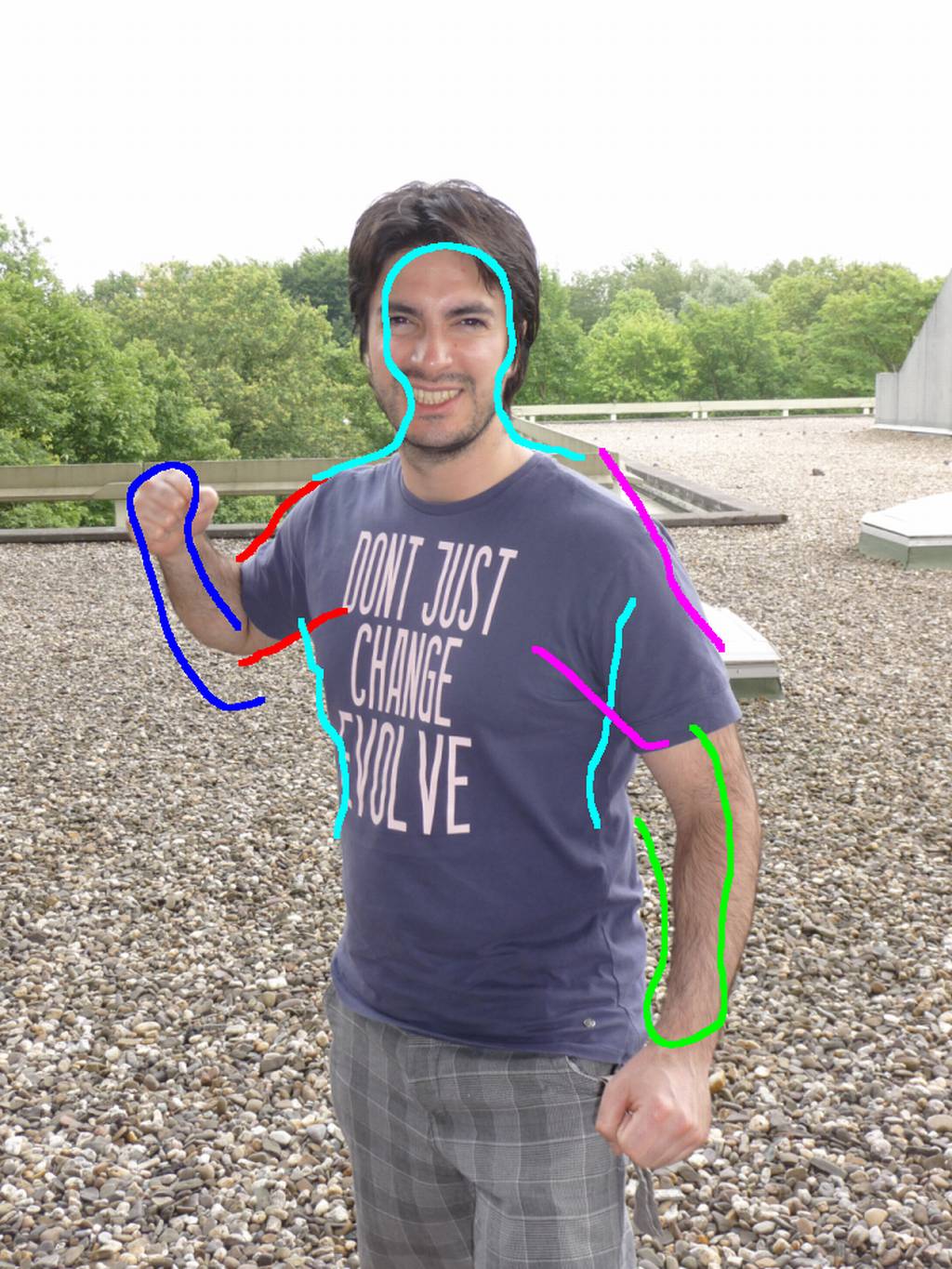}}
\hspace{0.1cm}
\subfloat[\label{fig:INIPUREResultsK_imgD}]{\includegraphics[width=0.45\textwidth]{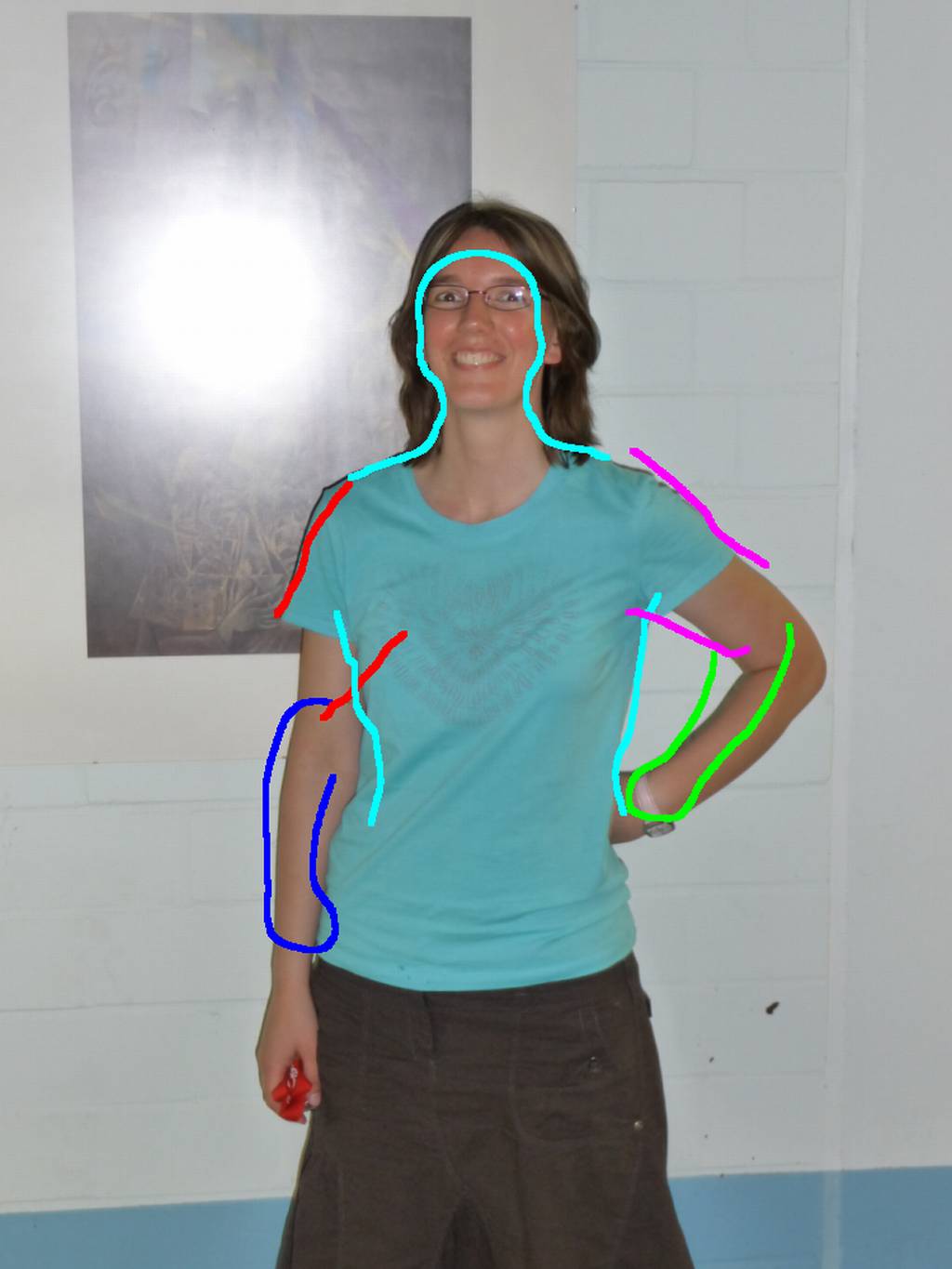}}
	\caption[Posture estimation results (sheet K)]{Posture estimation results (sheet K)}
	\label{fig:INIPUREResultsPlateK}
\end{figure*}

\modified{
To test the posture inference (Fig.~\ref{fig:SystemOverview}q)
capabilities of the proposed meta model `in the wild', we have recorded the
\emph{INIPURE} (Institut f\"{u}r NeuroInformatik -- PostURe
Estimation) database. This image set contains 63 upper body shots of
adult subjects of different genders and age
groups. Inter-subject variance in physique and worn attire is
significant; a subject's body pose may be severely foreshortened,
background clutter and scene illumination are relatively unconstrained.
For evaluation of matching, all images in the database have been
manually annotated with ground-truth 2D posture information.
All images (together with the system performance) are shown in figures~\ref{fig:INIPUREResultsPlateA} through~\ref{fig:INIPUREResultsPlateK}.
Raw images for comparison with other systems are available from the authors on request.
}
This allows to compare our
system's matching performance with human intuition: let
$\text{DT}_{b,i}(\mathbf{x})$ be a \emph{distance transformation map}
that stores the minimum Euclidean distance of any pixel in an INIPURE
\emph{benchmark} image $b$ (henceforth represented by
$\mathbf{I}_b(\mathbf{x})$) to body part $i$'s manually labeled
perimeter. Further, be $\mathbf{s}^*_{u,i}$ a binarized and thinned
representation of meta model $u$'s $i$'th shape template, projected to
$\mathbf{I}_b(\mathbf{x})$ according to our system's optimal posture
estimate. Nonzero pixels $\mathbf{h}_{u,i}\in\mathbf{s}^*_{u,i}$ shall
be provided in $\mathbf{I}_b(\mathbf{x})$'s coordinate system. With
that, the untruncated chamfer distance between $\mathbf{s}^*_{u,i}$
and the annotated perimeter of body part $i$ in benchmark image $b$
becomes~\citep{Shotton08a}
\begin{equation}	
d_{u,i,b}=\frac{1}{\sum{\mathbf{s}^*_{u,i}}}\sum_{\mathbf{h}_{u,i}\in \mathbf{s}^*_{u,i}}\text{DT}_{b,i}\left(\mathbf{h}_{u,i}\right)\,,
\label{eq:ChamferEvaluation}
\end{equation}
where $\sum{\mathbf{s}^*_{u,i}}$ is the total number of nonzero pixels
in $\mathbf{s}^*_{u,i}$.  Eq.~(\ref{eq:ChamferEvaluation}) allows to set
up the \emph{limb-averaged} model registration error $E_{u,b}$ 
as the average of all $N_\text{G}$ $d_{u,i,b}$.

\subsection{Reviewing the meta model}

Based on $E_{u,b}$, the quality and robustness of our meta model can
be assessed: in real-world settings, the succession of
sequence-specific body models integrated into the meta model can
hardly be controlled. Consequentially, the meta model's matching
performance has to be invariant against changes in model integration
order. To check for such invariance, basic permutation tests suffice:
a \emph{permutation table} with $N_\text{T}=50$ rows and $N_\text{M}$
columns is set up with each row $r$ containing a random permutation of
all $N_\text{M}$ body model indices.  Processing the table row by row,
sequence-specific body models are aggregated according to the row
entries, forming interim meta models
$\mathbf{M}_{\text{meta}_r}$. These models are then matched to 6
benchmark images selected from the INPURE database. To indicate good
matching performance, both mean registration error $\mu_{\text{P},b}$
and standard deviation $\sigma_{\text{P},b}$ should stay as low as
possible for all benchmark images; in particular, small standard
deviations allow to deduce that model integration order does not
significantly impact quality of posture inference.  Cols.~2 and 3 of
table~\ref{tbl:PermutationAndRepeatedLoadResults} shows that this
requirement is fulfilled, even for complex outdoor scenarios with
severely cluttered background.

A follow-up benchmark test aims at assessing the system's invariance
against \emph{redundant information}: given that the meta model
assembly routines operate properly and extract all accessible
information from the sequence-specific models, adding identical models
multiple times should not significantly alter matching
performance. Probing this hypothesis is straightforward: as model
integration order was experimentally shown to be meaningless, a fixed,
\emph{canonical} sequence of input models is selected first.

\begin{table}[t]
\centering
	\begin{tabular}{|l || c|c|}
	\hline
		Active switches & $\mu_\mathcal{C}$ (pixel) & $\sigma_\mathcal{C}$ (pixel) \\
		\hline
		$\tilde{S}$ & 171 & 113 \\
		\hline
		$\tilde{S},\tilde{G}$ & 62 & 71 \\
		\hline
		$\tilde{S},\tilde{G},\tilde{R}$ & 27 & 19 \\
		\hline
		$\tilde{S},\tilde{G},\tilde{R},\tilde{C}$ & 23 & 19 \\
		\hline
		$\tilde{S},\tilde{G},\tilde{R},\tilde{C},\tilde{M}$ & 14 & 10 \\
		\hline
		$\tilde{S},\tilde{G},\tilde{R},\tilde{C},\tilde{M},\tilde{A}$ & 11 & 3 \\
		\hline
	\end{tabular}
\caption{Mean and standard deviation of matching quality 
for different chosen cues.} \label{tbl:FeatureEvaluationResults}
\end{table}

Meta models $\mathbf{M}_{\text{meta}_m}$ are then learned by
integrating the canonical model sequence $m$ times, where
$m\in[0;N_\text{C}-1]$, with $N_\text{C}=10$. Posture estimation using
each $\mathbf{M}_{\text{meta}_m}$ is performed on the above benchmark
images; mean registration error and standard deviation are evaluated,
and cols.~4 and 5 of table~\ref{tbl:PermutationAndRepeatedLoadResults}
shows that posture estimation results remain precise (small
$\mu_{\text{R},b}$'s) and stable (small $\sigma_{\text{R},b}$'s),
regardless of artificially enforced data redundancy.

\modified{
\begin{figure*}
\subfloat[]{\includegraphics[width=0.3\textwidth]{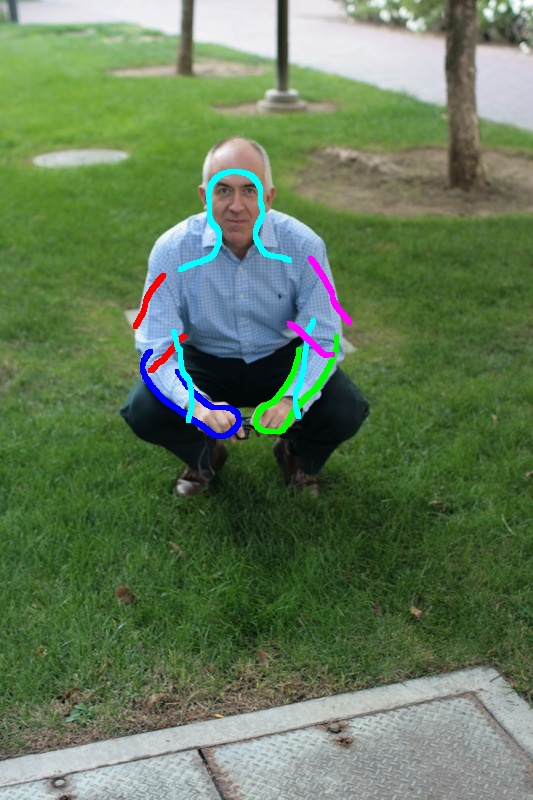}}
\hspace{0.1cm}
\subfloat[]{\includegraphics[width=0.3\textwidth]{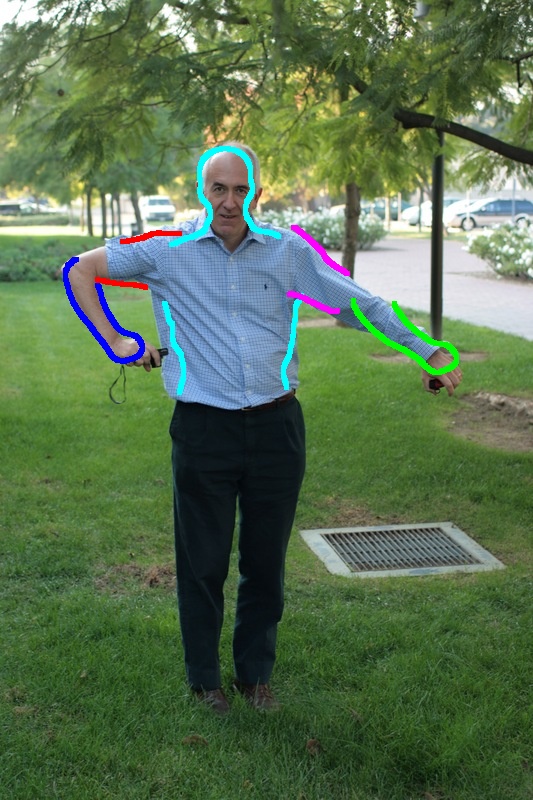}}
\hspace{0.1cm}
\subfloat[]{\includegraphics[width=0.3\textwidth]{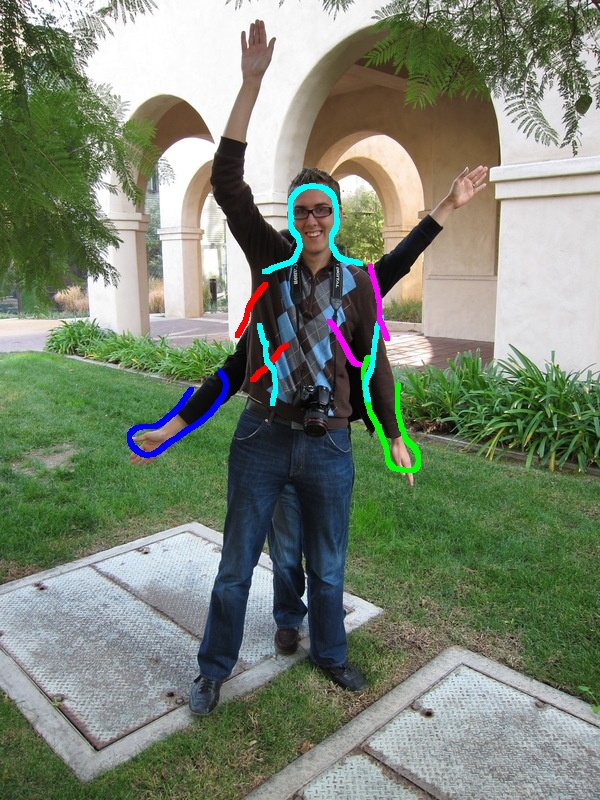}}
  \caption[]{\label{fig:PeronaResults}\modified{Some results for the Perona challenge \citep{Perona12}, see additional material for all of them.}}
\end{figure*}
}

\modified{
\subsection{Hyperparameter settings}
As both metamodel construction and matching are necessarily based on a
combination of several methods from the literature, the number of
hyperparameters from all these methods adds up, amounting to some 20
thresholds, template sizes, and other hyperparameters in this
work. Fortunately, these are not independent but can be adjusted
sequentially.  Working on the given training set, the hyperparameters
for background suppression, tracking and trajectory clustering were
adjusted one after the other to get reasonable behavior. None of them
are critical, no systematic optimization has been applied. The same
goes for the hyperparameters of shape, color and Gabor cues as well as
the combination parameters, which have been chosen using a subset of 6
images from the INIPURE dataset. These have been applied to the full
dataset. The Perona dataset has been evaluated without any
hyperparameter change. Of course, several hyperparameters imply
assumptions about ranges of scale, image resolution, etc., which might
need to be adjusted for new training data.}

\subsection{Experiments on real-world footage}

\begin{figure*}[t]
\centering	
\subfloat[\label{fig:NAOResultsA_imgA}]
{\includegraphics[width=0.32\textwidth]{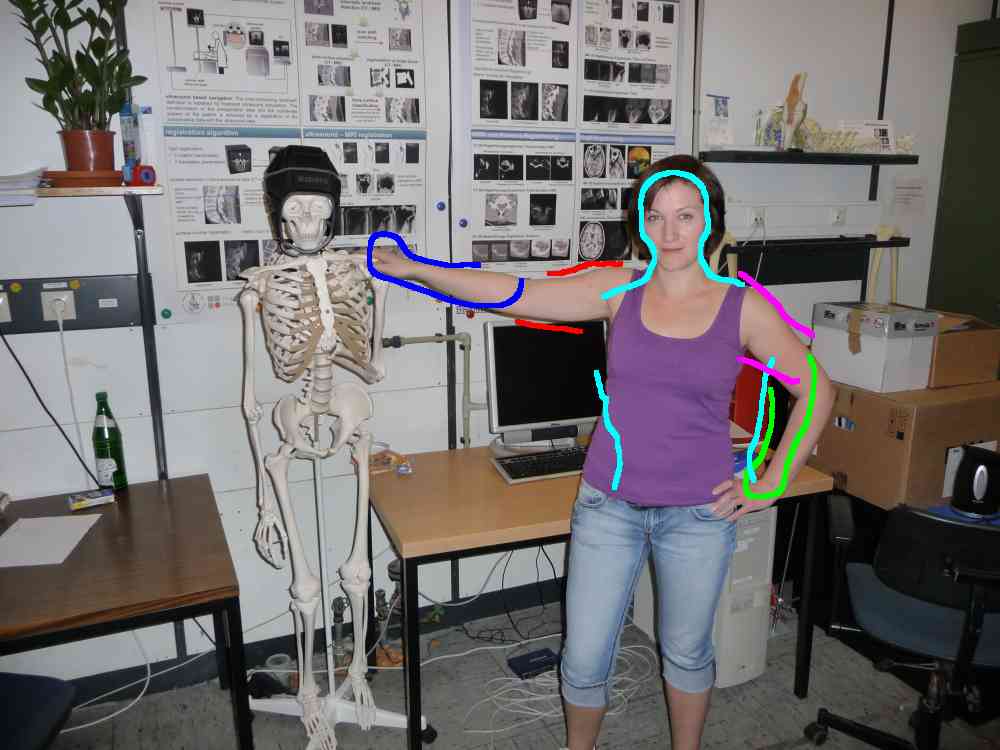}}
\hspace{0.1cm}
\subfloat[\label{fig:NAOResultsA_imgC}]	{\includegraphics[width=0.32\textwidth]{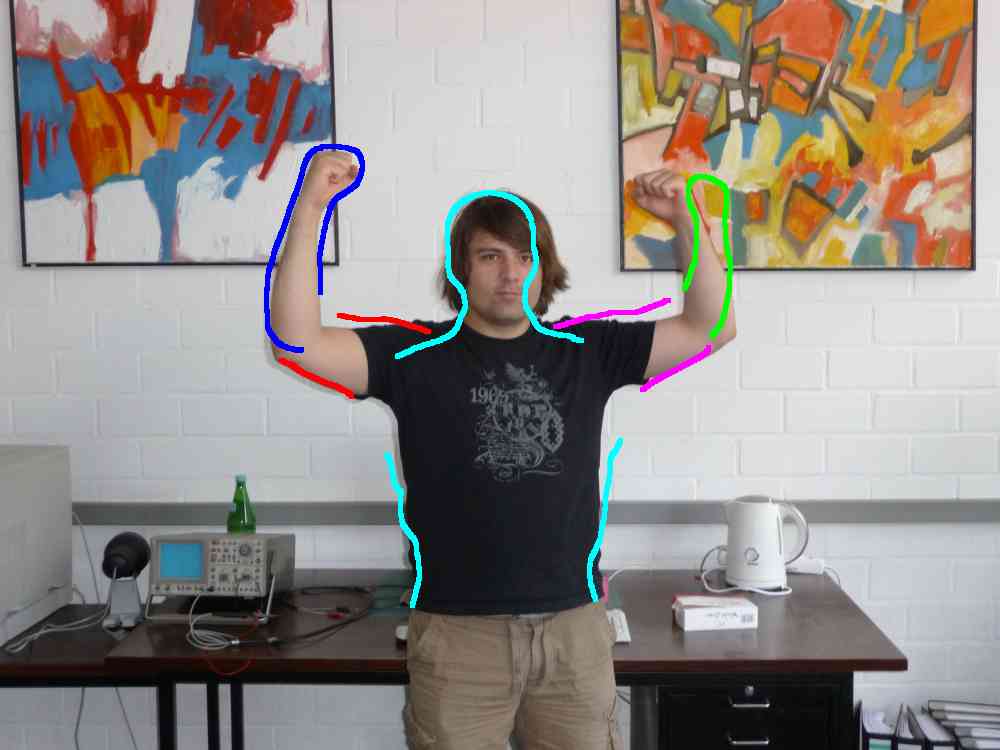}}
\hspace{0.1cm}
\subfloat[\label{fig:NAOResultsA_imgE}]
{\includegraphics[width=0.32\textwidth]{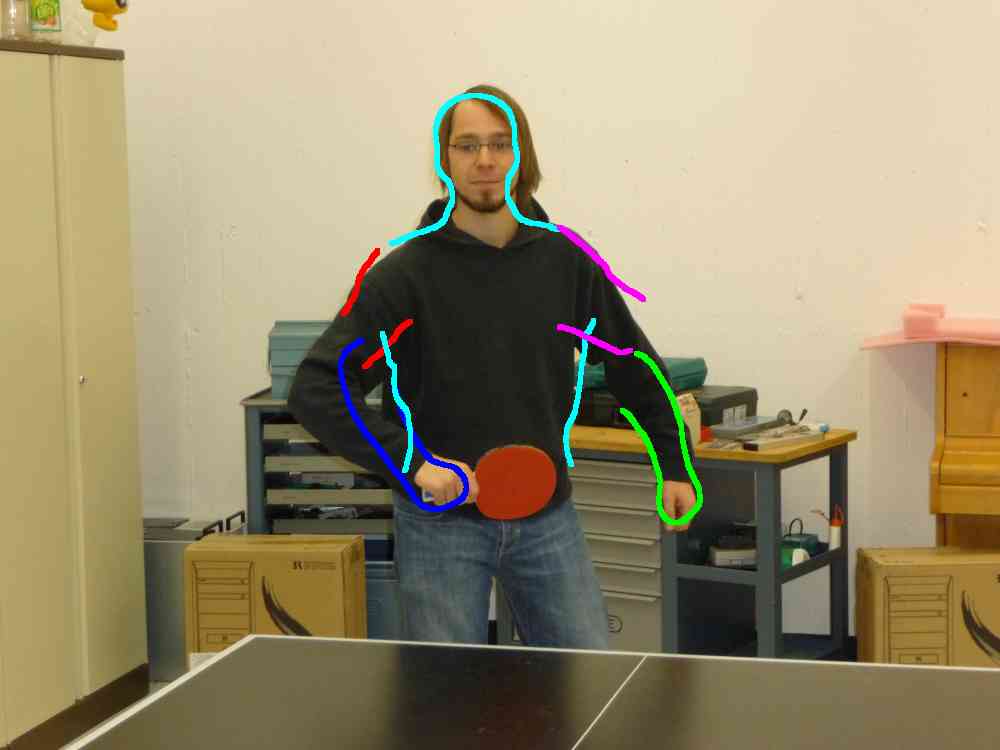}}
\\
\subfloat[\label{fig:NAOResultsA_imgB}]
{\includegraphics[width=0.32\textwidth]{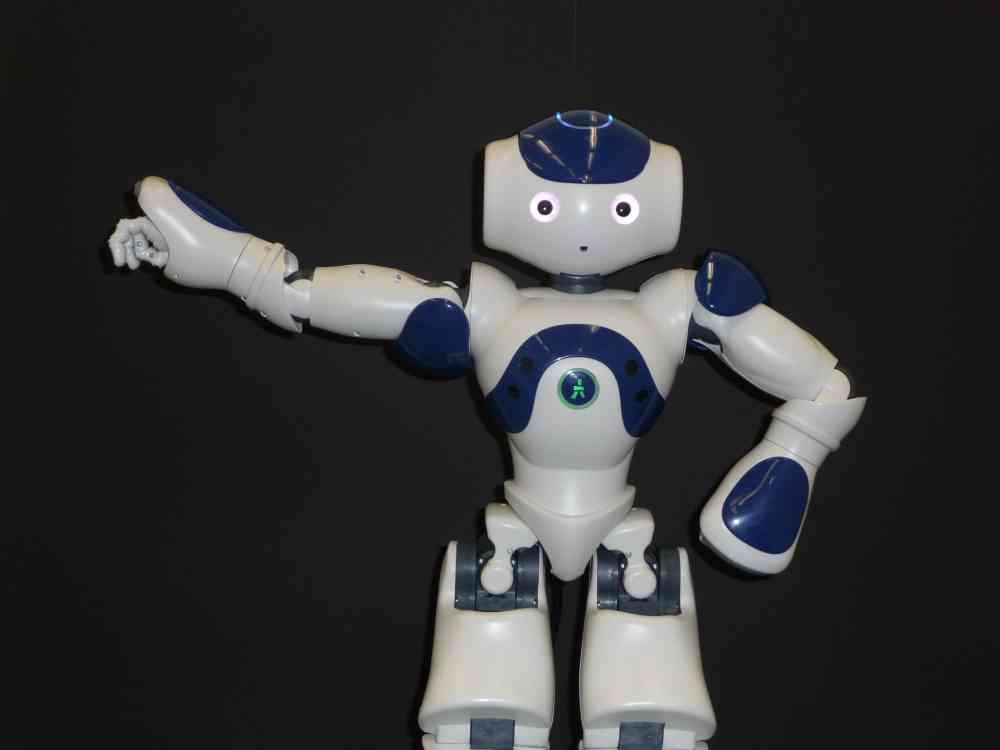}}
\hspace{0.1cm}
\subfloat[\label{fig:NAOResultsA_imgD}]
{\includegraphics[width=0.32\textwidth]{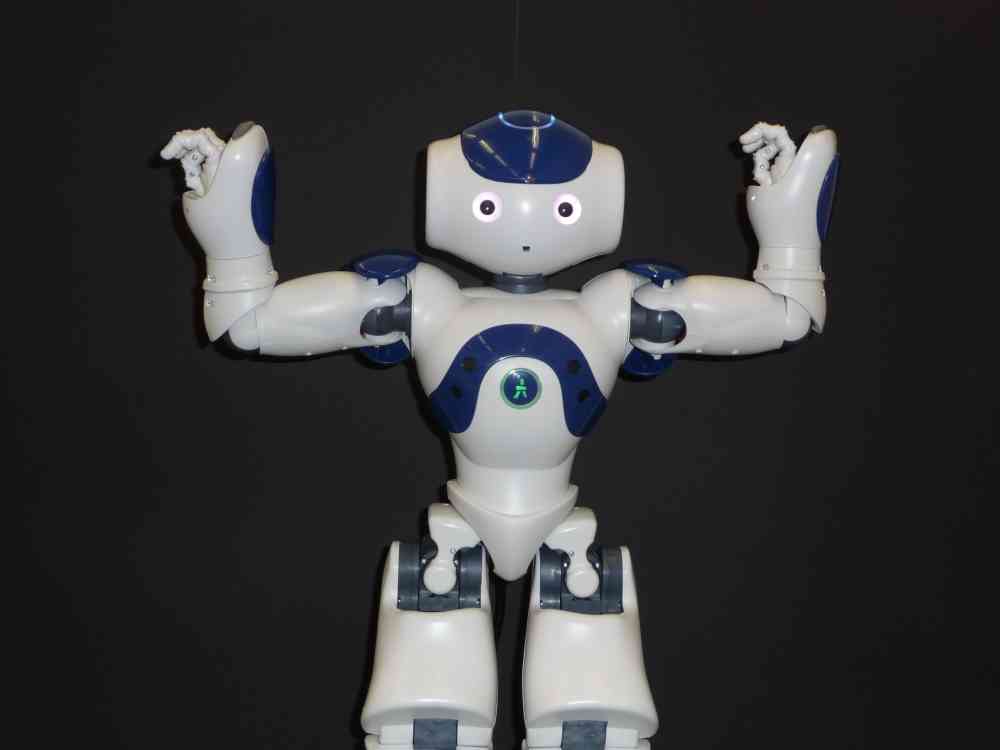}}
\hspace{0.1cm}
\subfloat[\label{fig:NAOResultsA_imgF}]
{\includegraphics[width=0.32\textwidth]{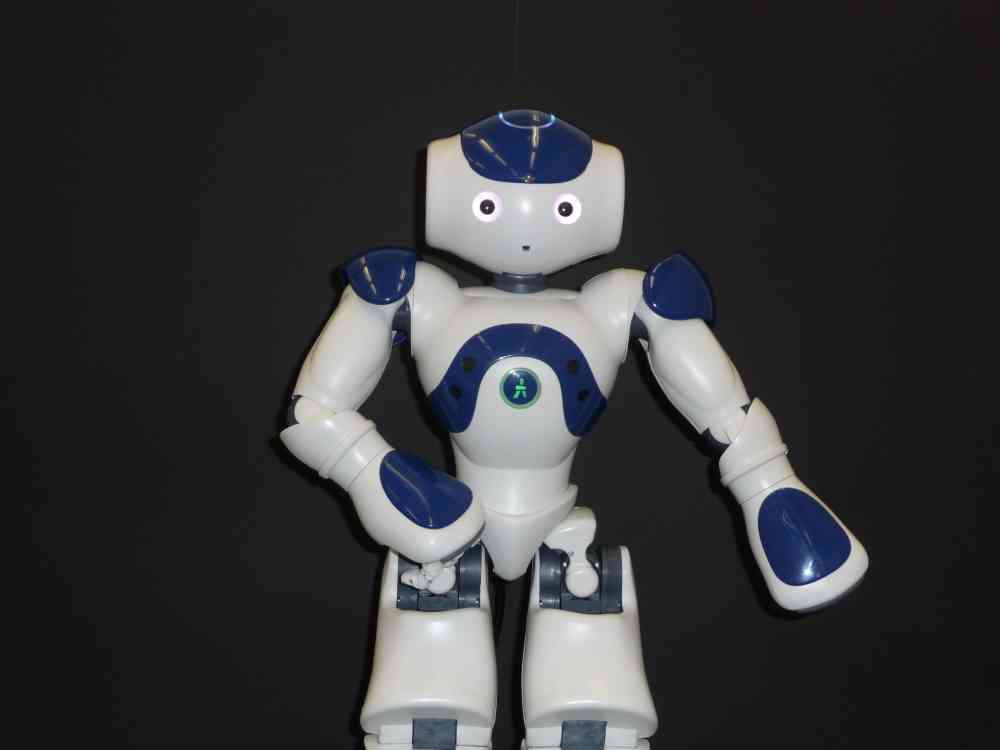}}

\caption[Posture mimicry performed by NAO]{Examples of successful posture
matching by the learned model (top row) and posture mimicry performed by the NAO robot.
(bottom row).}
	\label{fig:INIPUREResultsPlate}
\end{figure*}

\modified{ Stepping beyond these demonstrations of algorithmic
  soundness, benchmarking was extended to real-world imagery. We have
  applied the trained meta model with all cues to all 63 images in the
  INIPURE database. The results were inspected and the images divided
  in two categories, 17 complete failures and 46 acceptable
  matches. Some of the latter are depicted in
  Fig.~\ref{fig:INIPUREResultsPlate}, full results are provided as
  additional material.  The optimal posture estimate for the canonical
  meta model has been overlaid to the query images, qualitatively
  demonstrating that inferred posture comes close to human intuition.
}

\begin{figure*}[p]
	\centering	

\subfloat[\label{fig:INIPUREResultsB_imgA}]{\includegraphics[width=0.45\textwidth]{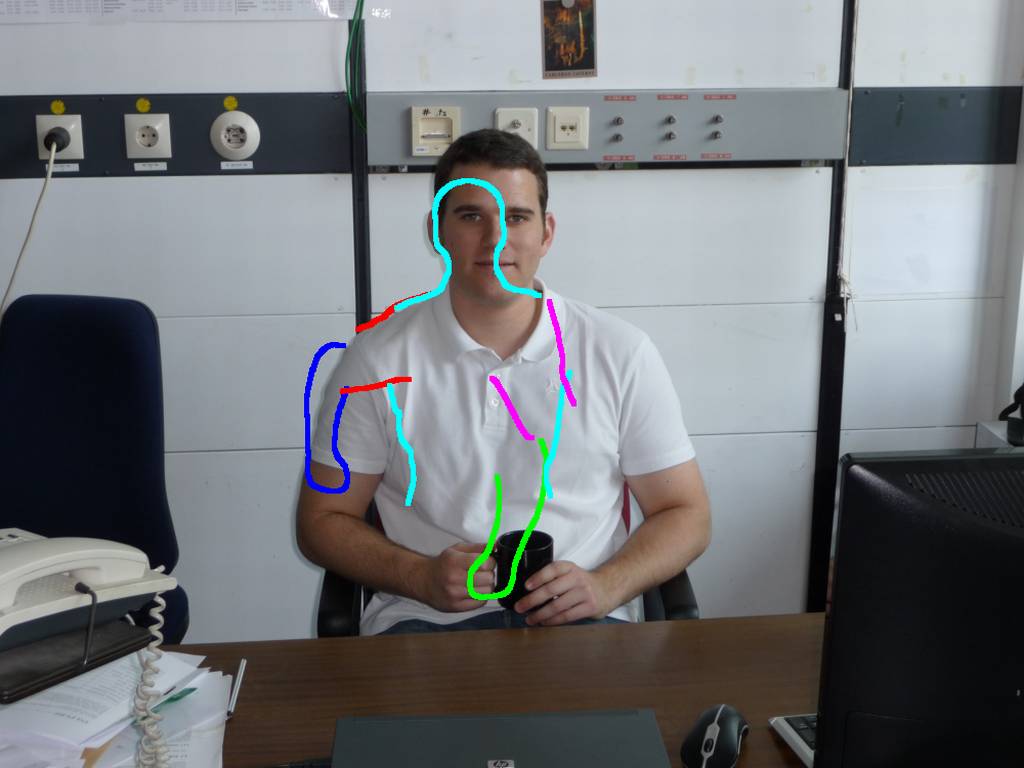}}
\hspace{0.1cm}
\subfloat[\label{fig:INIPUREResultsB_imgB}]{\includegraphics[width=0.45\textwidth]{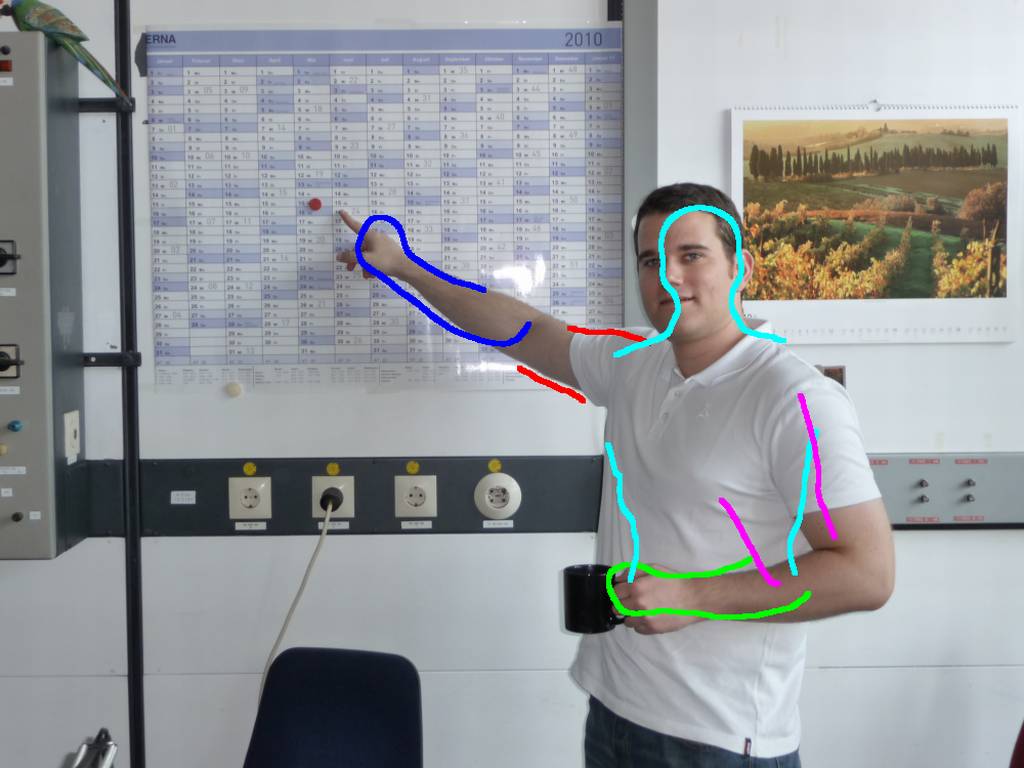}}\\

\subfloat[\label{fig:INIPUREResultsB_imgC}]{\includegraphics[width=0.45\textwidth]{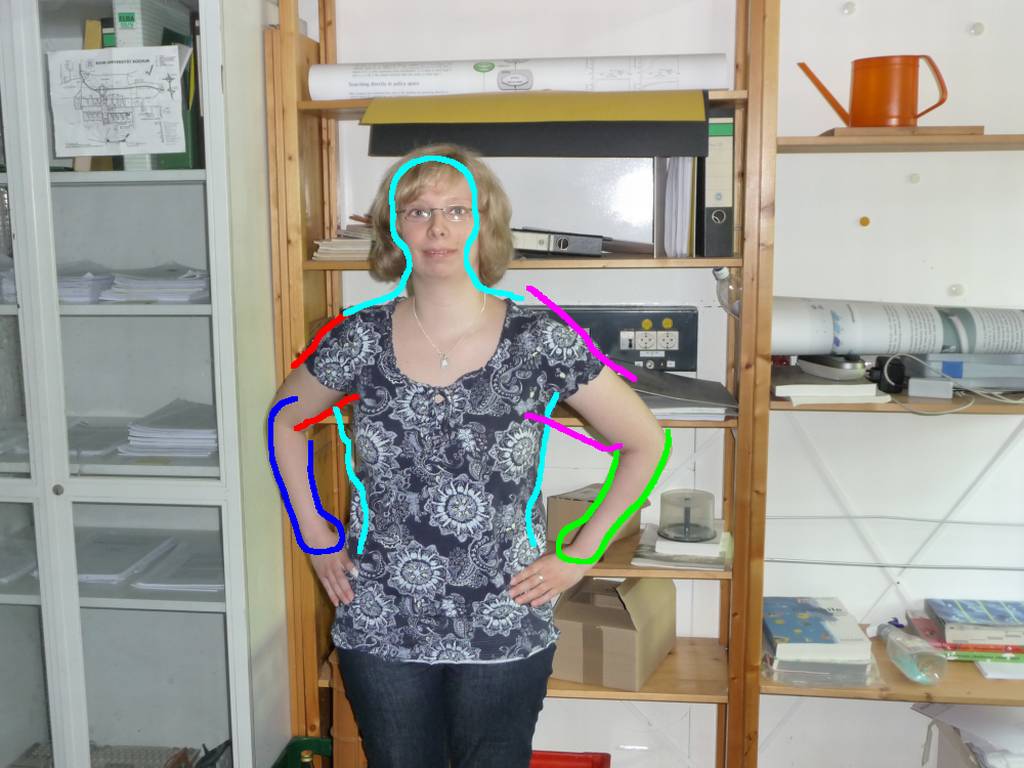}}
\hspace{0.1cm}
\subfloat[\label{fig:INIPUREResultsB_imgD}]{\includegraphics[width=0.45\textwidth]{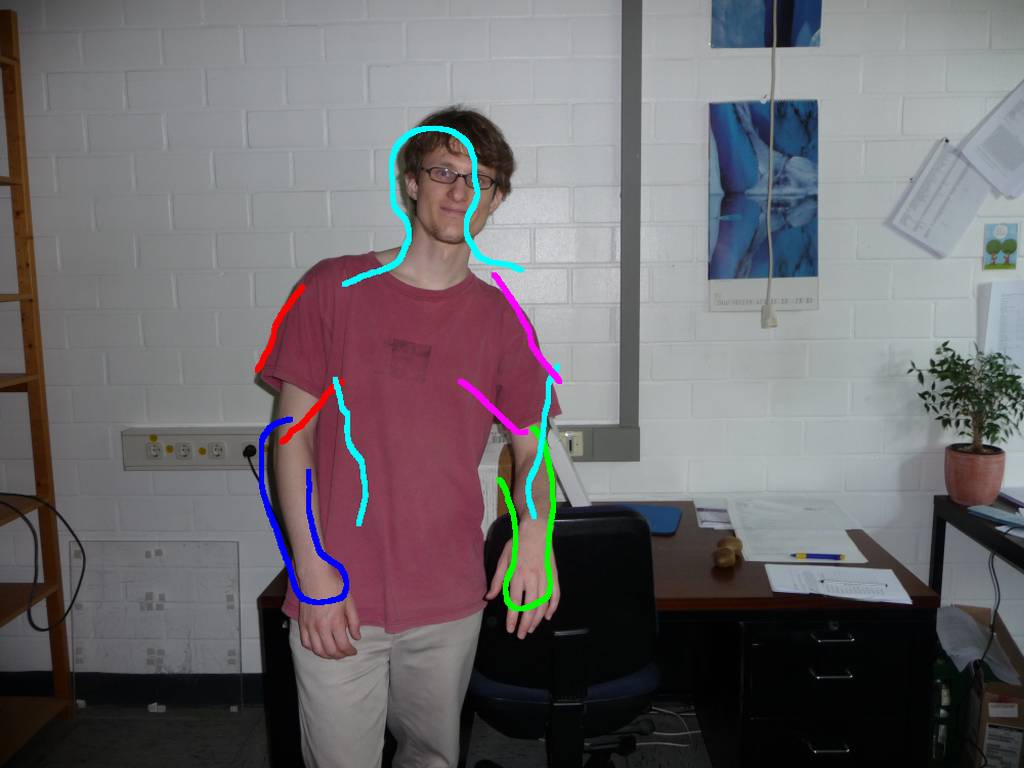}}\\

\subfloat[\label{fig:INIPUREResultsB_imgE}]{\includegraphics[width=0.45\textwidth]{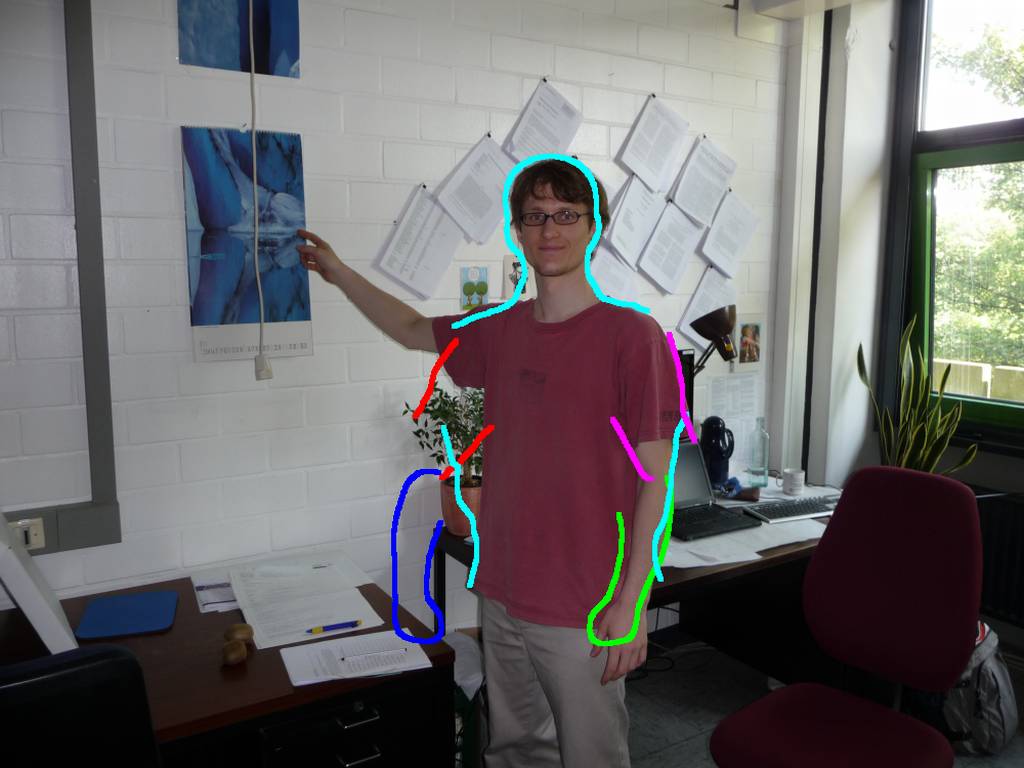}}
\hspace{0.1cm}
\subfloat[\label{fig:INIPUREResultsB_imgF}]{\includegraphics[width=0.45\textwidth]{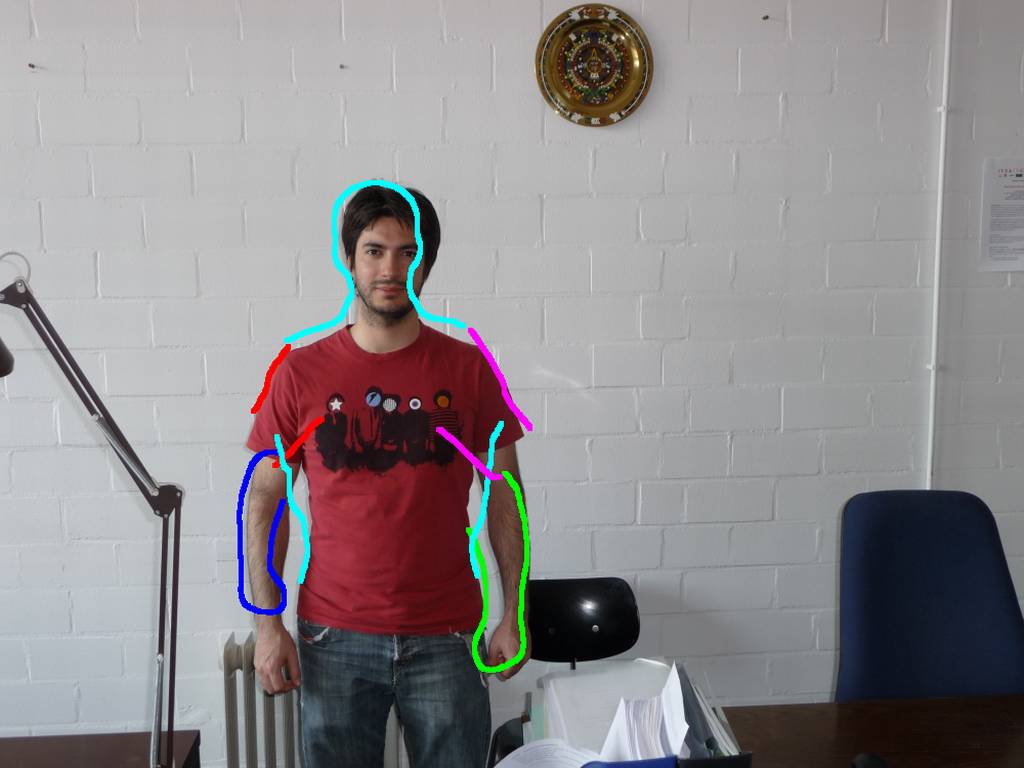}}\\

	\caption[Posture estimation results (sheet B)]{Posture estimation results (sheet B)}
	\label{fig:INIPUREResultsPlateB}
\end{figure*}

\modified{Increased visual complexity will
cause posture analysis to become less precise and might even evoke
partially wrong estimation results like in
figs.~\ref{fig:INIPUREResultsB_imgC} and~\ref{fig:INIPUREResultsB_imgD}.
}

\modified{With the acceptable matches, we further investigated the contributions
of the meta model's single cues to the matching success. To that end,
let meta model $\mathbf{M}_{\text{meta}_\mathcal{C}}$ be assembled
from the canonical input sequence introduced above. Further, assume
that use of shape and color features in
$\mathbf{M}_{\text{meta}_\mathcal{C}}$ can selectively be controlled
via `switches' $\tilde{S}$, resp. $\tilde{C}$. Other switches are
deemed available for Gabor-based limb pre-detection ($\tilde{G}$),
search space restriction by $\theta_{typ,(\cdot)}$ ($\tilde{R}$),
application of stimulus maps ($\tilde{M}$), and image enhancement by
chromatic adaptation ($\tilde{A}$), allowing to activate/deactivate
them on demand. The \emph{switch configuration} variable $\mathcal{C}$
shall represent the set of engaged switches; for instance,
$\mathcal{C}=\{\tilde{S},\tilde{G},\tilde{R},\tilde{C},\tilde{M},\tilde{A},\}$
indicates that $\mathbf{M}_{\text{meta}_\mathcal{C}}$'s capabilities
are in full function, while $\mathcal{C}=\{\}$ identifies an
inactivated meta model that becomes powerless in matching attempts.
}

As the following statistics serve for \emph{system intrinsic}
comparison, it is reasonable to defer overly complex outlier cases and
focus on the $N_{\text{B}}$ image subset during evaluation of the
average $\mu_{\mathcal{C}}$ and standard deviation
$\sigma_{\mathcal{C}}$ of $E_{\mathcal{C},b}$.
Table~\ref{tbl:FeatureEvaluationResults} shows that pure shape
information does not remotely suffice to ensure reliable posture
inference (row 1), as background clutter induces a large number of
false positive shape elements, causing imprecise (excessively large
mean) and unstable (large standard deviation) results. By adding
Gabor-based limb pre-detection routines (row 2), matching precision
increases, yet, as indicated by the large standard deviation, remains
unstable.

Locking body part rotation of all pre-detected meta limbs $i$ to
$\theta_{typ,i}$ (row 3) not only speeds up the matching process but
also allows to circumnavigate a range of false local
optima. Nevertheless, obtained results are still far from being useful
for reliable posture analysis. Exploiting color information (row 4)
further increases systemic performance. By activating the stimulus
maps (row 5), both mean error and standard deviation show another
sudden drop and registration quality increases
significantly. Switching on chromatic adaptation (row 6) additionally
boosts matching quality by allowing for more reliable color
analysis. Wrapping up, tbl.~\ref{tbl:FeatureEvaluationResults} clearly
demonstrates sophisticated cue fusion to be an inevitable mainstay of
successful meta model registration in significantly complex real-world
situations. Note that the above results were achieved using a
`well-behaved' image subset for system intrinsic testing. Stepping to
the \emph{unbiased} image set with $N_B$=63 images,
including all outlier cases, $\mu_\mathcal{C}$ becomes 25
pixels, while $\sigma_\mathcal{C}$ takes on a value of 30 pixels
(with all switches active).  

\begin{figure*}[p]
	\centering	
\subfloat[\label{fig:PeronaResultsA_imgA}]{\includegraphics[width=0.25\textwidth]{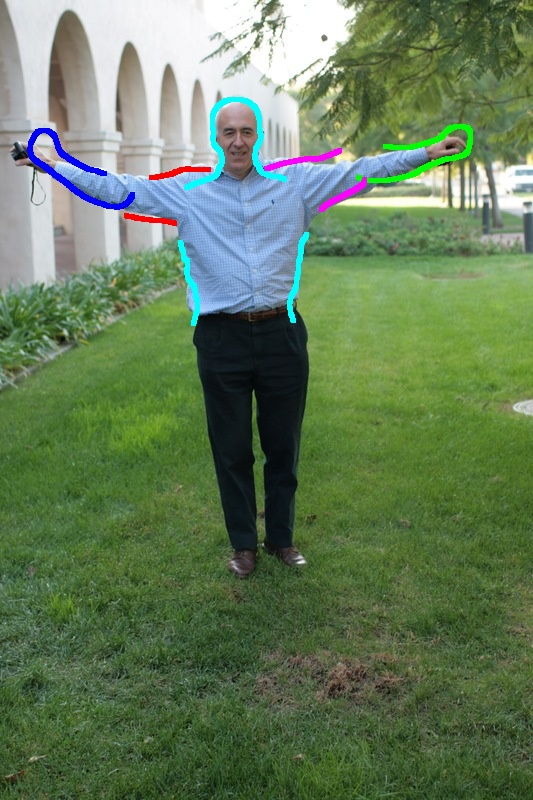}}
\hspace{0.1cm}
\subfloat[\label{fig:PeronaResultsA_imgB}]{\includegraphics[width=0.25\textwidth]{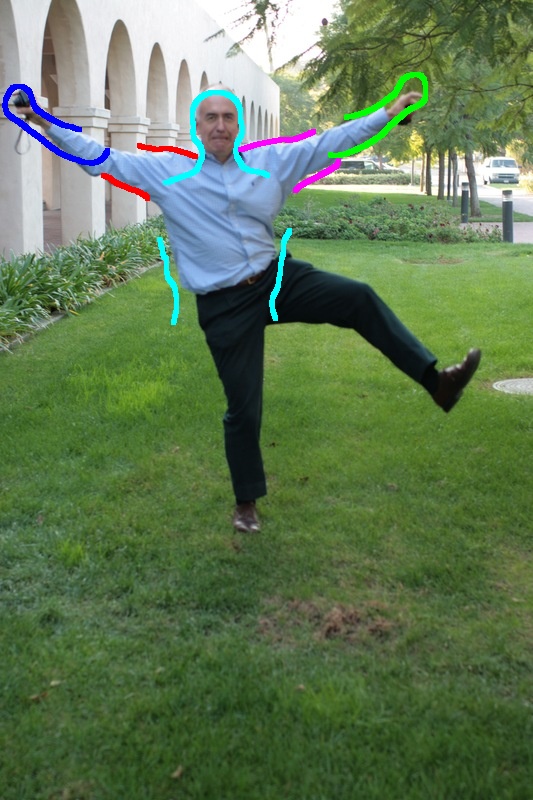}}
\hspace{0.1cm}
\subfloat[\label{fig:PeronaResultsA_imgC}]{\includegraphics[width=0.25\textwidth]{Images/fig_ImagesPerona/images-10.jpg}}\\
\subfloat[\label{fig:PeronaResultsA_imgD}]{\includegraphics[width=0.25\textwidth]{Images/fig_ImagesPerona/images-11.jpg}}
\hspace{0.1cm}
\subfloat[\label{fig:PeronaResultsA_imgE}]{\includegraphics[width=0.25\textwidth]{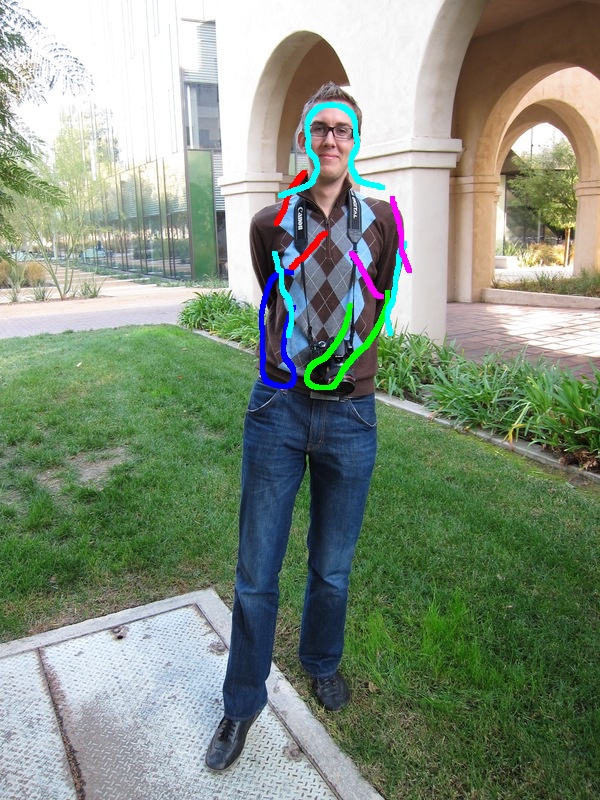}}
\hspace{0.1cm}
\subfloat[\label{fig:PeronaResultsA_imgF}]{\includegraphics[width=0.25\textwidth]{Images/fig_ImagesPerona/images-31.jpg}}\\
\subfloat[\label{fig:PeronaResultsA_imgG}]{\includegraphics[width=0.45\textwidth]{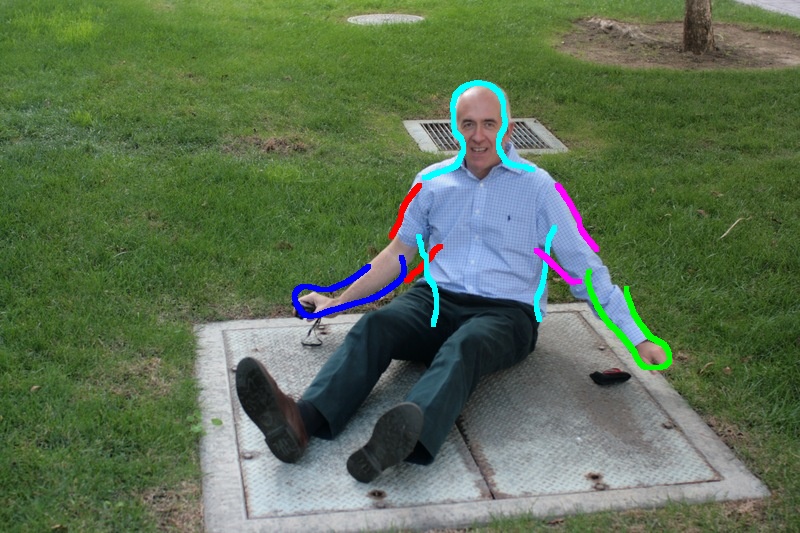}}
\hspace{0.1cm}
\subfloat[\label{fig:PeronaResultsA_imgH}]{\includegraphics[width=0.45\textwidth]{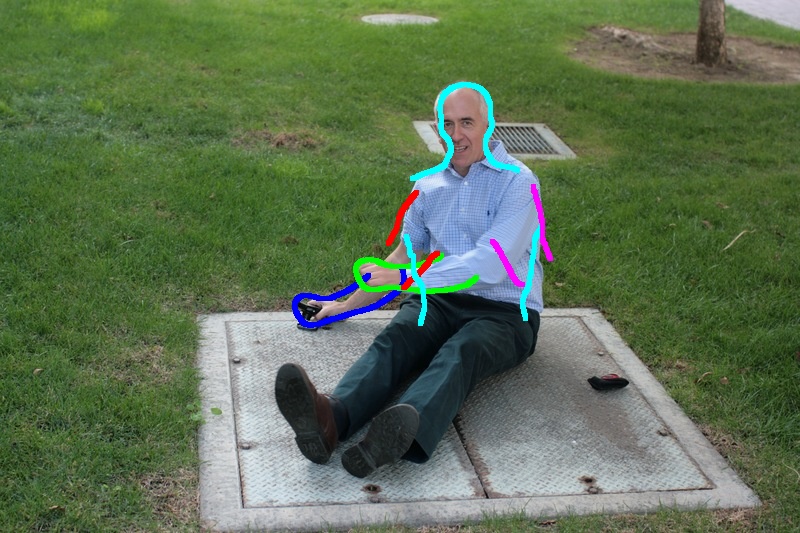}}
	\caption[Posture estimation results Perona (sheet A)]{Posture estimation results, `Perona November 2009 Challenge', by courtesy of Pietro Perona, (sheet A)}
	\label{fig:PeronaResultsPlateA}
\end{figure*}

\begin{figure*}[t]
	\centering	
\subfloat[\label{fig:PeronaResultsD_imgA}]{\includegraphics[width=0.45\textwidth]{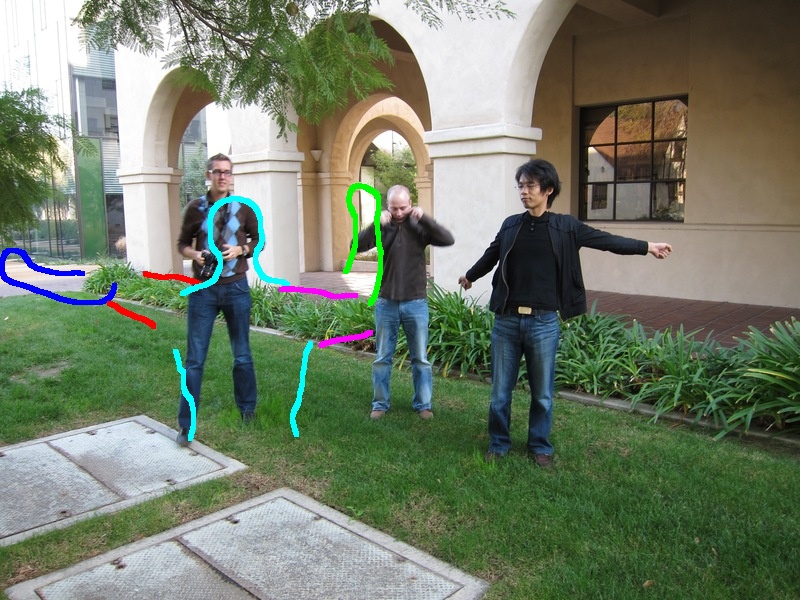}}
\hspace{0.5cm}
\subfloat[\label{fig:PeronaResultsD_imgB}]{\includegraphics[width=0.25\textwidth]{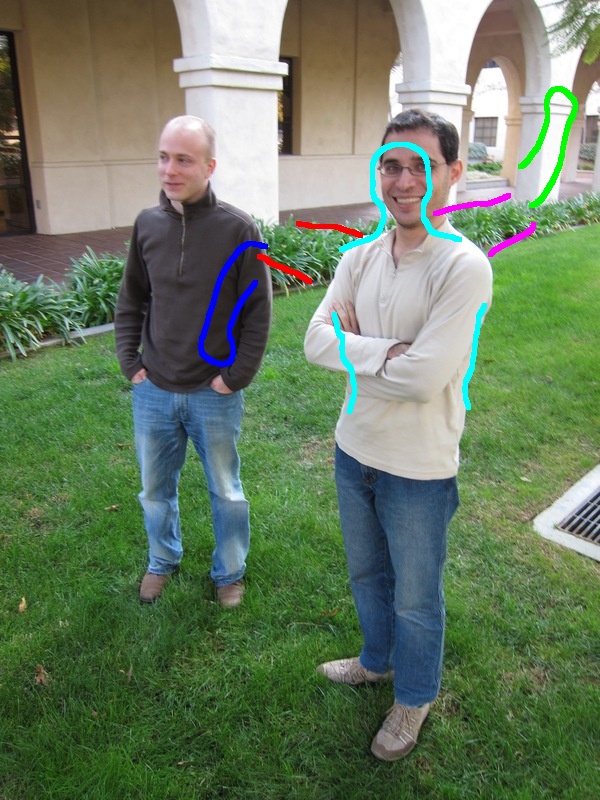}}\\
	\caption[Posture estimation results Perona (sheet D)]{Posture estimation results, `Perona November 2009 Challenge', by courtesy of Pietro Perona, (sheet D)}
	\label{fig:PeronaResultsPlateD}
\end{figure*}

To get a more comparative overview of the performance of our solution
in publicly available `real-world' scenarios, we tested our routines
on the `Perona November 2009 Challenge' \citep{Perona12}
dataset. While performing posture estimation on this less complex
image ensemble, poses were correctly recognized on 10 images out of
32, yielding $R_{ok}=31.25$. Here $R_{ok}$ defines the percentage of
`correctly' recognized poses by means of human intuition.  While this
measure is no true quantitative value, it allows to check the
estimation performance of our system on external image benchmarks
without the necessity for extensive pose tagging. As our system learns
from a single subject and has to generalize from autonomously acquired
knowledge (compared to massive training found in standard posture
estimation approaches) this value seems quite acceptable. \modified{ Some
matches are shown in~\ref{fig:PeronaResults}, full results in the
additional material, figs.~\ref{fig:PeronaResultsPlateA}
through~\ref{fig:PeronaResultsPlateD}.  For the \emph{
  INIPURE} dataset $R_{ok}$ becomes equal to $69.84$.}

\subsection{Timing considerations}
Calculations were done on an AMD
Phenom\texttrademark\ II X4 965, 3.4 GHz unit with 4 GB of RAM,
and an  NVIDIA\TReg\
GeForce\TReg9800 GT graphics adapter. 

\begin{table}[t]
\centering
	\begin{tabular}{|c||c|c|c|}
	\hline
		Model & frames & seconds total &seconds per frame\\
		\hline
		0&200&42&0.21 \\
		\hline
		1&200&46&0.23 \\
		\hline
		2&200&58&0.29 \\
		\hline
		3&600&80&0.13 \\
		\hline
		4&185&42&0.22 \\
		\hline
		5&109&29&0.26 \\
		\hline
		6&270&60&0.22 \\
		\hline
		7&232&56&0.24\\
		\hline
		8&207&49&0.24 \\
		\hline
		9&193&50&0.26 \\
		\hline
		10&249&58&0.23 \\
		\hline
		11&331&77&0.23 \\
		\hline
		12&248&53&0.21 \\
		\hline
		13&379&75&0.19\\
		\hline
		14&115&28&0.24 \\
		\hline
		15&75&\modified{21}&0.27 \\
		\hline
		16&460&83&0.18 \\
		\hline
		 \multicolumn{3}{c|}{} & $\oslash=$0.23\\
		 \cline{4-4}
	\end{tabular}
\caption{Construction times for sequence-specific models} \label{tbl:TimingResults}
\end{table}

Tbl.~\ref{tbl:TimingResults} shows processing times (third column) are
logged for each sequence-specific model from initial motion
segmentation to final PS generation. Normalization by the
number of frames in each sequence yields comparable
\emph{per-frame figures} (fourth column), whose mean of $\leq 0.25$
seconds per frame indicates that sequence-specific upper body models can
be learned reasonably fast.

With that, it becomes interesting to analyze meta model construction
timings; necessary figures are logged while performing above
permutation tests: be $T(\mathbf{M}_{\text{meta}_r})$ the time (in
seconds) it takes to assemble $\mathbf{M}_{\text{meta}_r}$ from row
$r$ of the permutation table. As meta model construction periods
linearly depend on the number of integrated, sequence-specific models,
the mean of all observed $T(\mathbf{M}_{\text{meta}_r})$,
$r=0,\ldots,N_\text{T}-1$ is divided by $N_\text{M}$, such that
$T_{\text{meta}}$ expresses the average meta model construction time
per integrated body pattern. Given the above hardware/software
configuration, values of $T_{\text{meta}}\approx$ 45 seconds per
integrated model are achieved, allowing to assemble meta models
swiftly, even for growing $N_\text{M}$, as meta model construction
time is linear in the number of integrated models. Eventually, assume
that meta model registration timings have been logged for all
`non-outlier' experiments. Finding the mean of the recorded figures,
the meta model matching cycle can be stated to consume 178 seconds per
image. This value could be improved with additional GPUs.

\modified{
  \begin{figure}
\includegraphics[width=\hsize]{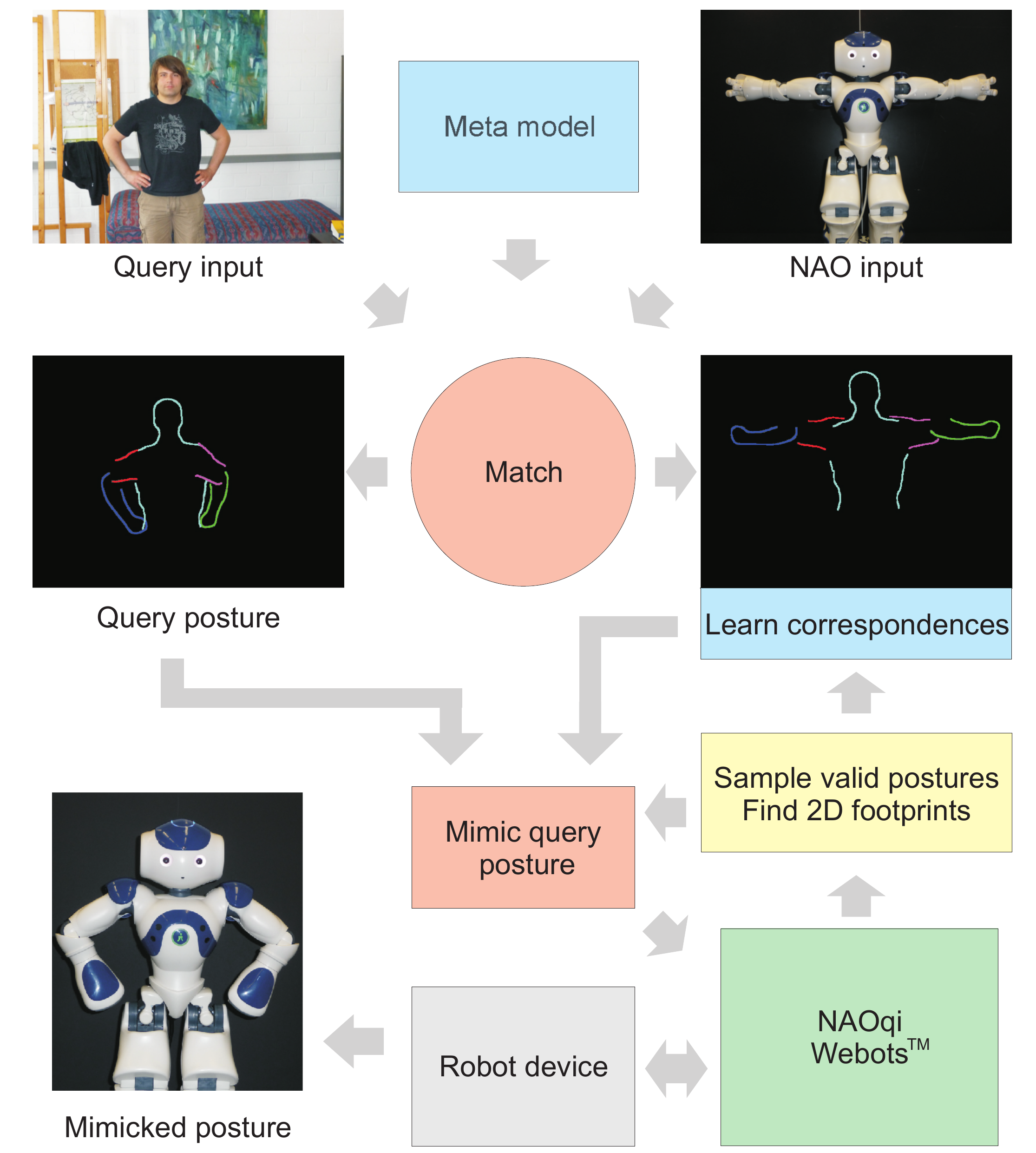}
  \caption[]{\label{fig:NaoMimicingCycle} NAO's mimicking behavior}
  \end{figure}
}

\subsection{Controlling the NAO robot device}

\modified{For the VBMC framework proposed in this paper, our experiments
demonstrate good generalization performance in weakly constrained
scenarios. To demonstrate another use, we have implemented
a behavior on a NAO robot device (manufactured by \emph{Aldebaran
 Robotics}), which uses the learned meta model to compare
an externally observed posture to its own posture and then making
appropriate movements to assume the same posture, thus mimicking
a human example.
}

\modified{Using  the \emph{NAOqi framework}, the
 robot's kinematic configuration can be controlled. From NAOqi's
 access to NAO's kinematics, an \emph{upper body skeleton} can be
 assembled that reliably traces the robot's posture changes. Assuming
 fronto-parallel motion patterns under orthographic projection, the
 single `bones' of the skeleton are expected to move parallel to the
 image plane, relative to which the rotation axes of the connecting joints
 should remain perpendicular. Therefore, the 3D
 skeleton is projected to the image plane by dropping depth
 information. The resulting \emph{skeletal footprint} $F(\mathbf{P})$
 yields a good 2D approximation of NAO's effective 3D pose
 $\mathbf{P}$, given that foreshortening is kept at bay (ensured
 manually).  Each projection is  augmented with the
 parameters of the 3D configuration it approximates; this way, NAOqi's
 actuator control routines can later be invoked to restore 3D posture
 corresponding to any particular skeletal footprint. With these
 preparations, NAO is guided manually through a range of
 fronto-parallel, collision-free upper body poses that are deemed
 typical for human beings. Footprints for each of the trained poses
 are sampled and stored, yielding an ample \emph{footprint repository}
 $\mathcal{R}$.
}

\modified{Now, if NAO is actuated to take on `T-like' posture
$\mathbf{P}_\text{T}$; a `snapshot' of this scenario is provided by
$\mathbf{I}_\text{T}(\mathbf{x})$. $\mathbf{M}_{\text{meta}}$ is
matched to $\mathbf{I}_\text{T}(\mathbf{x})$; all cues other than
shape are turned off in this registration process, realizing that
NAO's appearance (w.\,r.\,t.\ color and texture) does not even
remotely resemble the human look encoded in
$\mathbf{M}_{\text{meta}}$. Thus, stepping beyond shape features would
only hamper reliable model registration here. Following successful
matching, structural correspondences between the matched meta model
and the skeleton encoded by $F(\mathbf{P}_\text{T})$ are
learned. Assuming persistence of the established correspondences
allows to find, for any meta model configuration $\mathcal{L}$, the
most similar skeletal footprint in $\mathcal{R}$.
}

\modified{To actually mimic posture observed in an image
$\mathbf{I}_\text{Q}(\mathbf{x})$, the above relations come in handy:
let $\mathcal{L}^*$ represent the globally optimal posture inferred by
matching $\mathbf{M}_{\text{meta}}$ to
$\mathbf{I}_\text{Q}(\mathbf{x})$. Let further skeletal footprint
$F^*(\mathbf{P}^*)\in\mathcal{R}$ be the one most similar to
$\mathcal{L}^*$. The observed upper body pose is then replicated by
sending parameters encoded in $\mathbf{P}^*$ to NAOqi, which in turn
actuates NAO as to take on the desired kinematic configuration. For a
graphical overview of the described posture mimicking cycle, see
fig. \ref{fig:NaoMimicingCycle}.
}

\modified{Due to the discrete nature of footprints collected in $\mathcal{R}$,
posture mimicking results will necessarily be an approximation to
$\mathcal{L}^*$. In addition, NAO's hardware limits the spectrum of
natural upper body poses that can be emulated (for instance, folding
the robot's arms seems physically impossible). 
Notwithstanding these limitations, Fig.~\ref {fig:INIPUREResultsPlate}
qualitatively demonstrate the proposed framework's effectiveness in
mirroring posture observed in real-world footage.
}

\modified{
\section{Comparison with supervised approaches}
}

\modified{
We also challenged our system with publicly available
  benchmarks like the `Buffy Stickmen' \citep{Ferrari12} image
  ensemble. Here, recognition rates were negligible compared to
  contemporary posture estimators including, for instance,
  \citep{Eichner12}. In fact, our system occasionally found correct
  postures, yet these data didn't suffice to set up meaningful
  statistics. However, such behavior was expected: images in the
  `Buffy' set are extremely complex, showing strong illumination,
  pose, and scale variations. As our system had to learn posture
  estimation from scratch without human guidance, it can by no means
  compete with contemporary pose estimation solutions like
  \citep{Eichner12} who make intense use of manual training and inject
  significant a priori knowledge into their system (for instance, by
  utilizing pre-made upper body detectors).
}

\modified{
The current state of the art in human pose estimation in still images
is defined by~\cite{deeppose-2014}. The system consists of a body
detector, which preprocesses all images. Afterwards, a cascade of
convolutional neural networks is trained on 11000 images from
the FLIC~\cite{sapp-modec-2013} and Leeds~\cite{pose-leeds-2011}
datasets. Each of these images is manually annotated with the position
of 10 upper body joints. Data augmentation by mirroring  and
crop box variation yields a total of 12 million training images.
}

\modified{
Like in all feedforward neural network approaches, evaluation of a
single image is extremely fast at 0.1s on a 12 core CPU. Training is
more laborious and takes some 24 days on 100 ``workers'' (presumably
combined CPU/GPU machines). Success rates are around 80\% for arms.
The corresponding numbers for our system are 4253 video frames,
none of which is annotated for training. Training time is 16 minutes
on a PC, and evaluation of a single still image takes about 3 minutes.
It is successful in 70\% of the cases on our own dataset, and 31\% on the
Perona challenge. 
}

\modified{
Clearly, our system cannot match the performance of supervised systems
but it demonstrates a reasonable learning success in a different setting
with much fewer ressources. Furthermore, enhancing our matching with 
preprocessing by a body detector would certainly 
yield much better results, especially on the multiple person images. 
The goal of our experimentation was to show the capabilities and 
shortcomings of a body model learned in an unsupervised way.
}

\section{Conclusion}

The promising results above allow to state that OC principles might
well be used to alleviate the obstructive need for human supervision
that plagues conventional HPE/HMA solutions. Our system learns
conceptual representations of the upper human body in a completely
autonomous manner, the experiments show that the resulting meta model
achieves perceptually acceptable posture inference in moderately
complex scenes.  Nevertheless, much work remains to be done: one
idea would be to replace our motion segmentation scheme; switching to
methods found, e.\,g., in~\citep{Kumar08} could allow for system
training in scenarios of increased complexity. Beyond that,
parallelization techniques show potential in boosting model matching;
avoiding registration jitter in the learning stage would probably
result in improved Gabor models for the single meta limbs.

\subsection*{Acknowledgments}

This work was funded by the DFG in the priority program ``Organic
Computing'' (MA 697/5-1, WU 314/5-2, WU 314/5-3). We thank our
colleagues at the Institut f\"ur Neuroinformatik for posing for the
testing data.
  

\section{References}
\bibliographystyle{elsarticle-harv}
\bibliography{references}

\end{document}